\documentclass{article}

\ifdefined\pdfminorversion\pdfminorversion=7\fi

\let\PaperAddContentsLine\addcontentsline
\usepackage{iclr2027_conference,times}
\let\addcontentsline\PaperAddContentsLine

\usepackage{amsmath,amsfonts,bm}

\def\eqref#1{equation~\ref{#1}}

\def\1{\bm{1}}

\DeclareMathAlphabet{\mathsfit}{\encodingdefault}{\sfdefault}{m}{sl}
\SetMathAlphabet{\mathsfit}{bold}{\encodingdefault}{\sfdefault}{bx}{n}

\IfFileExists{ICLR-booktabs.sty}{\usepackage{ICLR-booktabs}}{\usepackage{booktabs}}
\usepackage{amssymb}
\usepackage{pifont}
\usepackage{graphicx}
\usepackage{float}
\usepackage{capt-of}
\usepackage{enumitem}
\usepackage{etoc}
\usepackage{etoolbox}
\usepackage{hyperref}
\usepackage{multirow}
\usepackage{tabularx}
\usepackage{wrapfig}
\usepackage{needspace}
\usepackage{xurl}
\usepackage{xcolor}
\usepackage{colortbl}

\definecolor{tablegray}{gray}{0.94}

\NewCommandCopy{\PaperSelectFont}{\selectfont}
\usepackage{CJKutf8}
\NewCommandCopy{\CaptionCJKSelectFont}{\selectfont}
\RenewCommandCopy{\selectfont}{\PaperSelectFont}

\makeatletter
\renewcommand{\paragraph}{\@startsection{paragraph}{4}{\z@}
    {1.5ex plus .5ex minus .2ex}{-1sp}{\normalfont\normalsize\bfseries}}

\newcommand{\graytablerow}{\noalign{\begingroup\color{tablegray}
    \hrule width\linewidth height\dimexpr\ht\@arstrutbox+\dp\@arstrutbox\relax
    \vskip-\dimexpr\ht\@arstrutbox+\dp\@arstrutbox\relax\endgroup}}
\makeatother
\definecolor{supportgreen}{HTML}{217A4B}
\definecolor{supportamber}{HTML}{B7791F}
\definecolor{supportred}{HTML}{C13B3A}
\newcommand{\fullsupport}{\textcolor{supportgreen}{\ding{51}}}
\newcommand{\partialsupport}{\textcolor{supportamber}{\ding{108}}}
\newcommand{\nosupport}{\textcolor{supportred}{\ding{55}}}

\newcommand{\gainpoints}[1]{\textcolor{supportgreen}{\ensuremath{\mathbf{+#1}}}}
\newcommand{\losspoints}[1]{\textcolor{supportred}{\ensuremath{\mathbf{-#1}}}}

\definecolor{evidenceink}{HTML}{254B70}
\definecolor{evidenceblue}{HTML}{EAF2FA}
\definecolor{evidencelight}{HTML}{F5F8FC}
\definecolor{evidenceline}{HTML}{CBD8E5}
\definecolor{evidencegreen}{HTML}{167A70}
\definecolor{evidencemuted}{HTML}{687787}
\definecolor{evidenceorange}{HTML}{E9A36B}

\hypersetup{
    colorlinks=true,
    citecolor=blue,
    linkcolor=blue,
    urlcolor=blue,
    linktoc=all,
    bookmarksdepth=2
}

\newcommand{\institutionlogo}[1]{
    \raisebox{-0.28\height}{\includegraphics[height=0.15in]{#1}}
}

\title{EvoCowork: A Benchmark for\\
Real-World Office Tasks}

\author{
    \textbf{Zhou Liu}\textsuperscript{1}\thanks{Equal contribution.},\,
    \textbf{Xingyan Chen}\textsuperscript{2,*},\,
    \textbf{ZhaoYang Han}\textsuperscript{1},\,
    \textbf{Zewei Pan}\textsuperscript{3},\\[-1pt]
    \textbf{Zeli Su}\textsuperscript{4},\,
    \textbf{Ningxin Pan}\textsuperscript{1},\,
    \textbf{Xue Hu}\textsuperscript{5},\,
    \textbf{Kaixiong Ting}\textsuperscript{6},\,
    \textbf{Wentao Zhang}\textsuperscript{1}\thanks{Corresponding author.}\\[3pt]
    \normalfont\small
    \begin{tabular}{@{}c@{\quad}c@{\quad}c@{}}
        \textsuperscript{1}\institutionlogo{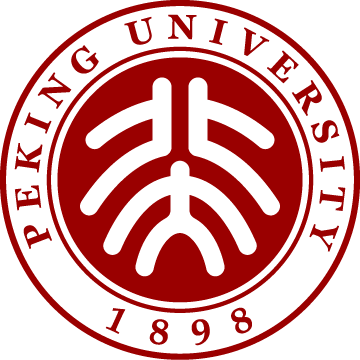}\,Peking University
        & \textsuperscript{2}\institutionlogo{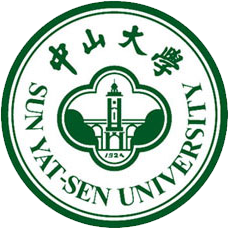}\,Sun Yat-sen University
        & \textsuperscript{3}\institutionlogo{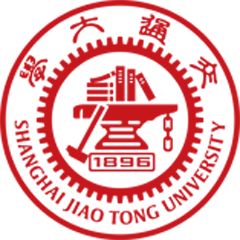}\,Shanghai Jiao Tong University\\[2pt]
        \textsuperscript{4}\institutionlogo{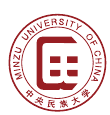}\,Minzu University of China
        & \textsuperscript{5}\institutionlogo{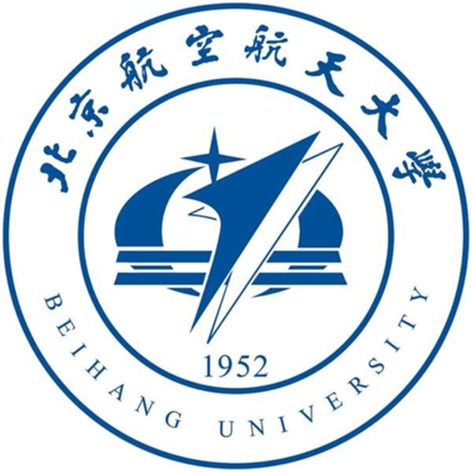}\,Beihang University
        & \textsuperscript{6}\institutionlogo{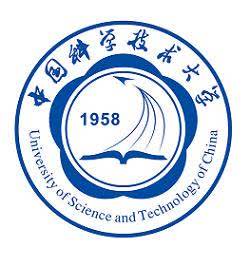}\,University of Science and Technology of China
    \end{tabular}
}

\iclrfinalcopy

\begin{document}
\raggedbottom
\etocdepthtag.toc{main}
\maketitle
\fancyhead{}
\renewcommand{\headrulewidth}{0pt}

\begin{abstract}
Office workflows are a demanding testbed for self-improving agents because
their deliverables must be correct, natively editable, and mutually consistent.
A central question is whether gains on feedback tasks reflect reusable procedures
rather than task-specific fixes. We introduce \textsc{EvoCowork}, a 240-task
benchmark for evaluating harness-level recursive self-improvement across four
office modalities, with model weights held fixed. The benchmark covers
presentations, documents, spreadsheets, and cross-office workflows, and scores
submitted artifacts for content, native structure, formulas, cross-file
consistency, and visual quality where needed. Its protocol separates feedback,
revision selection, and final testing: methods learn from single-modality tasks,
then are evaluated on unseen single-modality tasks and cross-office workflows
under a shared initial harness and explicit budgets. In static evaluations of
two models and four harnesses, the best configuration scores 42.25 overall and
30.62 on cross-office tasks. Starting from the same MSA harness with GLM~5.3,
ACE and AHE improve in-domain scores by 4.90 and 2.64 points, respectively,
but change cross-office scores by only $+0.03$ and $-1.72$. Thus, gains on the
file types that provide feedback do not necessarily transfer to coordinated
workflows. \textsc{EvoCowork} makes this distinction explicit while recording
the capability, transfer, and cost of harness evolution. Code and task packages:
\href{https://github.com/DeepThinkingZhouLiu/cowork-evolution-benchmark}{GitHub repository}.
\end{abstract}

\begin{figure}[H]
    \centering
    \includegraphics[width=0.8\linewidth]{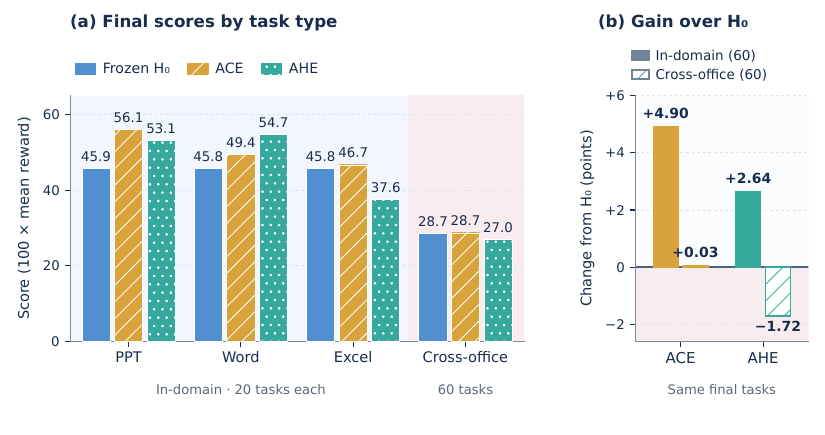}
    \caption{Final scores and evolution gains with GLM~5.3 and a shared MSA seed.
    (a)~Frozen $H_0$, ACE, and AHE on 20 tasks per single-modality track and
    60 cross-office tasks. (b)~Changes from $H_0$ on the 60 in-domain tasks and
    60 cross-office tasks, matching Table~\ref{tab:evolution-main-results}.
    Both methods improve the in-domain average, while cross-office gains are
    near zero for ACE and negative for AHE. Bars summarize the fixed final task
    sets from one run per method.}
    \label{fig:motivating-diagnostic}
\end{figure}

\section{Introduction}

Office agents often carry out related work: they update a workbook, use its figures
in a report, and prepare a presentation from the same material. Professional
workflow benchmarks test this work across files and applications
\citep{vidgen2026apexagents,tang2026workspacebench}, and their feedback can guide
how an agent inspects files, uses tools, and checks outputs. Cross-office
workflows are the sharpest instance: their deliverables must stay natively
structured and mutually consistent, which distinguishes a harness that learned
reusable procedures from one that merely fixed the tasks that generated the
feedback. These behaviors depend on the harness---the instructions, tools, context, and
execution logic around the model \citep{lee2026metaharnessendtoendoptimizationmodel}.
We study harness-level recursive self-improvement (RSI) by repeatedly revising this
layer while model weights stay fixed, and test whether the revisions make later
office work more reliable.

Existing benchmarks provide complementary ways to assess this capability. Office
benchmarks measure workflow execution and deliverable quality, including
requirements that span several files
\citep{wang2024officebench,zhou2026omegauseofficeval,tang2026workspacebench}.
Harness-evolution benchmarks test revisions on held-out tasks, including office
tasks, though without coordinated multi-file deliverables
\citep{huang2026evobench,wu2026harnessdevllmscreateevolve}, while FinEvo-Bench
measures learning across professional financial workflows
\citep{deng2026finevobenchlongitudinalbenchmarkselfevolving}. These evaluations
address both initial capability and improvement from experience. However, none of
these suites measures whether changes learned on one file type transfer to
workflows whose deliverables must agree, and none scores office artifacts at the
requirement level needed to see that difference. Our question concerns a specific
form of transfer: do changes learned from feedback on individual file types help on
new workflows whose deliverables must agree? Reporting these two conditions
separately matters because an overall evolution gain can hide improvement on one
and regression on the other.

Several design choices are needed to make this comparison informative. Feedback
tasks must be separated from final tests, because fixing the tasks that prompted a
revision does not establish that it will help elsewhere. A shared initial harness
and model make gains comparable, while explicit budgets help distinguish the
effect of an update from that of additional computation
\citep{wang2026rethinking,zheng2026seagymevaluationenvironmentselfevolving}.
Scores must check the requested work in the files themselves: a report can disagree
with its workbook, and a spreadsheet can contain the right number without the
required formula. Evaluation therefore needs checks of content, native structure,
and consistency, with visual assessment where layout matters. Rubric judgments
also need their own reliability checks
\citep{peng2026llmasajudgereliablyverifyrubrics}. Reserving cross-office workflows
for final testing then asks whether procedures learned on individual file types
remain useful when information must pass between files.

\textsc{EvoCowork} implements these requirements in a suite of 240 tasks adapted
from public sources, covering presentations, documents, spreadsheets, and
cross-office workflows. Each task has fixed input files and explicit output
requirements. Automated checks and, where applicable, visual rubrics score the
submitted artifacts and provide feedback. Evolution methods learn from
single-modality feedback tasks; selection tasks choose the revision, and the two
final partitions measure generalization to new single-modality tasks and transfer
to cross-office workflows; the selected harness is frozen before final testing. Each method starts from the same harness and model.
Search and inference costs are recorded alongside changes in task scores.

Our evaluation compares ACE \citep{zhang2025ace} and AHE \citep{lin2026ahe} with
the frozen initial harness, $H_0$. The static results in
Table~\ref{tab:static-main-results} describe the starting capabilities of the
evaluated agents. The paired comparison in Table~\ref{tab:evolution-main-results}
tests changes from $H_0$ on unseen tasks. ACE gains 2.46 points overall and AHE
0.46, but both split along the two test sets: in-domain scores rise by 4.90 and
2.64 points while cross-office scores change by $+0.03$ and $-1.72$. Selection
scores point the other way for AHE, which loses 7.03 points there and still gains
overall on the final tasks.

Overall, our contributions are threefold: (1) an artifact-aware Office benchmark
of 240 tasks whose checks score content, native editability, and cross-file
consistency in the submitted files; (2) a controlled RSI protocol that separates
feedback, selection, and final testing, and measures gains on unseen tasks and
transfer to cross-office workflows under a shared starting harness and explicit
budgets; and (3) an empirical finding that gains on isolated file types do not
automatically transfer to cross-office coordination, which reporting the two
conditions separately keeps visible.

\section{Related Work}

\subsection{Office Tasks and Artifact Evaluation}

Office-agent benchmarks cover both interactive workflows and work-product quality.
OSWorld evaluates real desktop interaction \citep{xie2024osworld}, while OfficeBench
emphasizes planning across office applications \citep{wang2024officebench}.
APEX-Agents tests professional work across applications
\citep{vidgen2026apexagents}, and GDPval uses expert comparisons of occupational
deliverables \citep{patwardhan2025gdpvalevaluatingaimodel}. Native office files
also require distinct forms of evaluation. Office Comprehension Benchmark tests
understanding of their content and structure \citep{shaik2026officecomprehensionbenchmark};
PPTArena and PPT-Eval test PowerPoint editing and creation
\citep{ofengenden2025pptarena,gandhi2026ppteval}, and SpreadsheetBench~2 covers
business spreadsheet workflows \citep{zhu2026spreadsheetbench2}.
OmegaUse-OfficeVal evaluates office-suite deliverables with economic grounding
\citep{zhou2026omegauseofficeval}, while Workspace-Bench tests dependencies among
heterogeneous files \citep{tang2026workspacebench}. These tasks motivate evaluating
the work product that a task requires, with scoring tied to its intended use
\citep{hua2026designingbenchmarksknowledgework}. \textsc{EvoCowork} uses native
office deliverables to test whether harness improvements transfer from individual
file types to coordinated workflows.

\subsection{Controlled Evaluation of Agent Harnesses}

ACE updates a context playbook \citep{zhang2025ace}, while AHE and Meta-Harness
edit harness code using execution evidence
\citep{lin2026ahe,lee2026metaharnessendtoendoptimizationmodel}.
Self-Harness checks revisions for regressions, while DarwinX searches a population
of harnesses with the model fixed
\citep{zhang2026selfharnessharnessesimprove,zhang2026darwinxevolvingagentharnesses}.
Evaluations must separate revision gains from extra computation
\citep{wang2026rethinking}.
HarnessOpt-Bench, Evo-Bench, and HarnessDev evaluate revisions on held-out tasks
\citep{ursekar2026harnessopt,huang2026evobench,wu2026harnessdevllmscreateevolve};
PAST-Bench and ContinualSkillBench study experience reuse
\citep{xue2026pastbenchbenchmarkingfoundationsrecursive,guan2026continualskillbenchllmagentstruly}.
S3Gym separates self-testing, self-judging, and self-improvement
\citep{shi2026s3gymllmsturnselftesting}, while Aspire studies evolution from vague
goals under hidden evaluation \citep{wu2026aspiremodelsselfevolvevague}.
SEAGym compares methods across training, validation, ID/OOD tests, and cost
\citep{zheng2026seagymevaluationenvironmentselfevolving}, while FinEvo-Bench
tracks learning through sequential financial tasks
\citep{deng2026finevobenchlongitudinalbenchmarkselfevolving}.
\textsc{EvoCowork} tests transfer from single-modality feedback to cross-office
work: ACE and AHE share a model and initial harness, and each selected revision
is frozen before testing native file structure and agreement across deliverables.

\suppressfloats[t]
\begin{figure}[t]
    \centering
    \includegraphics[width=\linewidth]{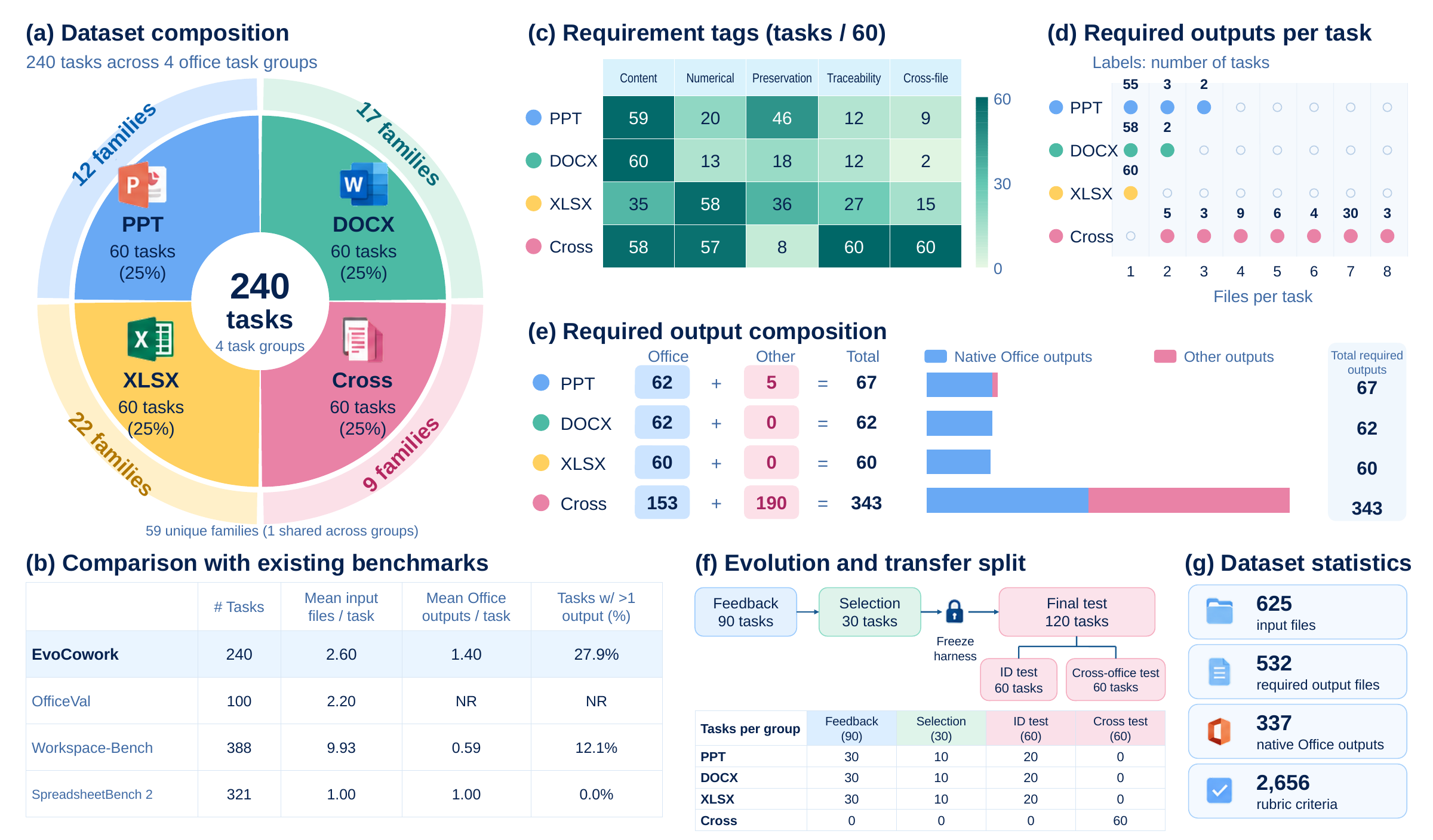}
    \caption{Task composition, artifact requirements, and evolution splits in
    \textsc{EvoCowork}. Panels (a--b) show task coverage and released-task metadata
    across benchmarks; (c--e) show requirement tags and output demands; (f--g) show
    the evaluation split and dataset totals. Requirement tags overlap. NR denotes
    an unavailable comparable field.}
    \label{fig:benchmark-overview}
\end{figure}

\section{The EvoCowork Benchmark}
\label{sec:benchmark}

\subsection{Design Rationale}
\label{sec:design-rationale}

Office tasks require both correct information and usable files. A report can state
the right numbers while omitting required editable tables; a workbook and its
summary can each look plausible while reporting different totals. We therefore
evaluate the submitted artifacts against explicit content, structure, and
consistency requirements. Visual checks supplement these requirements on tasks where layout matters.
This connects the work being represented, the required product, and the result
actually scored, as advocated in work-centered benchmark design
\citep{hua2026designingbenchmarksknowledgework}.

\begin{wraptable}[10]{r}{0.44\linewidth}
    \vspace{-0.4\baselineskip}
    \centering
    \caption{Per-modality scale and artifact counts.
    Families and sources are deduplicated.}
    \label{tab:benchmark-overview}
    \scriptsize
    \setlength{\tabcolsep}{4pt}
    \renewcommand{\arraystretch}{1.05}
    \begin{tabular}{lrrrr}
        \toprule
        Modality & Fam. & Src. & In & Out \\
        \midrule
        Presentation  & 12 & 6 & 105 &  67 \\
        Document      & 17 & 5 & 107 &  62 \\
        Spreadsheet   & 22 & 9 & 190 &  60 \\
        Cross-office  &  9 & 4 & 223 & 343 \\
        \midrule
        \textbf{Total} & \textbf{59} & \textbf{11} & \textbf{625} & \textbf{532} \\
        \bottomrule
    \end{tabular}
\end{wraptable}
The four modalities separate work within a file type from coordination across
types. Presentation, document, and spreadsheet tasks test their respective
operations, while cross-office tasks require agents to carry information through
dependent deliverables. With 60 tasks per modality, each contributes equally to
the static overall score. This balanced allocation defines the scope of the release.
Workflow categories describe coverage, whereas difficulty labels summarize the
performance of a fixed calibration panel. Their generalization must be tested
with a held-out model.

Public source tasks provide requests and inputs. Freezing those inputs
and specifying the output contract lets different harnesses face the same
task and supports replay after a harness changes. Static evaluation measures
initial task performance. Evolution evaluation asks whether changes learned from
feedback improve performance on separate final tasks. We therefore separate
feedback, revision selection, and final evaluation, and reserve cross-office
tasks to test transfer from single-modality feedback. Separate update-validation
and transfer views are also used by SEAGym
\citep{zheng2026seagymevaluationenvironmentselfevolving}; here the transfer
changes the coordination required among office deliverables
(Appendix~\ref{app:split-boundaries}).

\subsection{Task Scope and Taxonomy}

\textsc{EvoCowork} scores the files an agent submits. Tasks ask agents to create,
edit, analyze, repair, or audit office files. Task families describe operations
such as reconstruction, formula writing, visual redesign, and reconciliation
across files. Within each modality, we also group tasks by workflow structure.
Presentation tasks range from editing a slide to producing a complete deck from
several sources. Document tasks cover content editing, native-object
reconstruction, evidence synthesis, and controlled review. Spreadsheet tasks
cover local transformations, workbook computation, source reconciliation, and
decision workflows. Cross-office tasks range from handing off one file to
coordinating several dependent outputs; their 343 required outputs across 60
tasks reflect the coordination demands that make them harder than single-modality
work (Table~\ref{tab:benchmark-overview}). These categories describe coverage;
empirical difficulty is assigned separately.
Appendix~\ref{app:difficulty-taxonomies} gives the definitions and examples.

Difficulty is calibrated by one common protocol. Every modality is evaluated by the
same frozen panel $\mathcal{A}$ of model--harness configurations under matched budgets
and a common seed policy. For task $i$ with visual scoring enabled, harness $a$, and seed $s$, the task reward
and panel summary are
\begin{equation}
    r_{i,a,s}=0.7r^{\mathrm{det}}_{i,a,s}+0.3r^{\mathrm{vis}}_{i,a,s},
    \qquad
    \bar r_i=\frac{1}{|\mathcal{A}|S}
    \sum_{a\in\mathcal{A}}\sum_{s=1}^{S}r_{i,a,s},
    \qquad d_i=1-\bar r_i,
    \label{eq:empirical-difficulty}
\end{equation}
where $r^{\mathrm{det}}$ is the deterministic artifact score,
$r^{\mathrm{vis}}$ is the task-specific visual score, and $d_i$ is empirical
difficulty. Tasks without visual scoring use $r_{i,a,s}=r^{\mathrm{det}}_{i,a,s}$. Tasks are ranked
by $\bar r_i$ within each modality and assigned to precommitted 20/30/30/20 bins for
\emph{Easy}/\emph{Medium}/\emph{Hard}/\emph{Expert}; a boundary tie triggers an audit
against that modality's structural regime before stable task-ID tie-breaking.

Table~\ref{tab:benchmark-comparison} compares evaluation objects,
artifact-related capabilities, and evolution protocols. Raw task counts use each source's released evaluation unit.
\textsc{EvoCowork} focuses on source-grounded, editable Office deliverables with
isolated replay and task-specific artifact rewards.

\begin{table}[!t]
    \centering
    \caption{Office-agent benchmark capabilities.
    \fullsupport/\partialsupport/\nosupport{} indicate full, partial,
    and no support; P/D/S/X denote presentation, document, spreadsheet, and
    cross-office tasks.}
    \label{tab:benchmark-comparison}
    \scriptsize
    \resizebox{\linewidth}{!}{
    \begin{tabular}{@{}lrlcccccc@{}}
        \toprule
        Benchmark & Tasks & Scope & Real source & Editable & Isolated & Atomic & Three-way & Harness \\
                  &       &       & grounding & artifacts & replay & reward & split & mutation \\
        \midrule
        PPT-Eval~\citep{gandhi2026ppteval} & 120 & P
            & \partialsupport & \fullsupport & \partialsupport & \fullsupport & \nosupport & \nosupport \\
        OmegaUse-OfficeVal~\citep{zhou2026omegauseofficeval} & 100 & P/D/S
            & \fullsupport & \fullsupport & \partialsupport & \fullsupport & \nosupport & \nosupport \\
        OfficeBench~\citep{wang2024officebench} & 300 & Multi-app
            & \partialsupport & \partialsupport & \fullsupport & \partialsupport & \nosupport & \nosupport \\
        OSWorld~\citep{xie2024osworld} & 369 & Desktop
            & \fullsupport & \partialsupport & \fullsupport & \partialsupport & \nosupport & \nosupport \\
        SpreadsheetBench~\citep{ma2024spreadsheetbench} & 912 & S
            & \fullsupport & \fullsupport & \partialsupport & \partialsupport & \nosupport & \nosupport \\
        TheAgentCompany~\citep{xu2024theagentcompany} & 175 & Workplace
            & \partialsupport & \partialsupport & \fullsupport & \fullsupport & \nosupport & \nosupport \\
        \midrule
        STATE-Bench~\citep{microsoft2026statebench} & 450 & Enterprise
            & \nosupport & \nosupport & \fullsupport & \partialsupport & \partialsupport & \partialsupport \\
        HarnessOpt-Bench~\citep{ursekar2026harnessopt} & 4 targets & General
            & \partialsupport & \partialsupport & \fullsupport & \partialsupport & \fullsupport & \fullsupport \\
        Evo-Bench~\citep{huang2026evobench} & 608 & S/O/G
            & \partialsupport & \partialsupport & \fullsupport & \partialsupport & \partialsupport & \fullsupport \\
        \midrule
        \textbf{\textsc{EvoCowork} (ours)} & \textbf{240} & \textbf{P/D/S/X}
            & \fullsupport & \fullsupport & \fullsupport & \fullsupport & \fullsupport & \fullsupport \\
        \bottomrule
    \end{tabular}}
\end{table}

\subsection{Source-Grounded Task Construction}

We select tasks from public, license-compatible sources and adapt them
individually. The presentation track draws from PPTArena~\citep{ofengenden2025pptarena},
OmegaUse-OfficeVal~\citep{zhou2026omegauseofficeval},
\href{https://github.com/zapier/AutomationBench}{AutomationBench},
TheAgentCompany~\citep{xu2024theagentcompany},
\href{https://github.com/0717376/cowork_bench}{Cowork-Bench}, and
OfficeBench~\citep{wang2024officebench}; the cross-office track draws from
AutomationBench, Cowork-Bench, OfficeBench, and OmegaUse-OfficeVal.
Replaying an upstream instance requires more than copying its prompt. Some
requests leave the deliverable implicit or depend on mutable application and network
state. A broad end-state check may miss an incorrect formula, an uneditable object,
or a contradiction across deliverables. Each adaptation therefore freezes the
required inputs, maps the request into an explicit deliverable contract, records
checksums and the exact upstream revision, and rebuilds or strengthens the evaluator
where necessary. Shared sources, templates, and near-duplicate workflows receive
leakage-group identifiers for overlap audits and controlled partitioning.
The presentation, document, spreadsheet, and cross-office source inventories and
upstream task identifiers are reported in Appendix~\ref{app:source-records},
together with their frozen revisions and licenses.

To verify task quality, we sampled 24 tasks per modality, stratified by
difficulty tier and upstream source, and each task receives two independent
reviews from a three-person panel, with the third member adjudicating
disagreements (Appendix~\ref{app:task-review-interface}). Reviewers read the
task instruction, source inputs, reference answer, and rubric, and assess
whether the instruction is unambiguous, the reference answer factually correct,
and every stated requirement satisfied. Tasks flagged for ambiguity or incorrect
reference answers were revised before inclusion.

\subsection{Replayable Task Packages and Artifact-Aware Judges}

Each presentation or cross-office task runs as a self-contained package.
The agent receives the instruction, fixed input files, execution environment,
and output requirements. Task-specific deterministic
judges inspect file validity, editable structures, factual content, formulas,
cross-artifact consistency, provenance, and preservation constraints as applicable.
Tasks with visually judged PPTX or PDF deliverables additionally use a task-specific
rubric for hierarchy, spacing, balance, visual fidelity, and professional quality with
a fixed Codex--VLM evaluator. Exact-recreation tasks expose the reference rendering to
this evaluator; open-ended generation tasks do not. The two scores are retained
separately and combined using the 70/30 rule in
Equation~\ref{eq:empirical-difficulty}. Reference and negative-output replays audit the
deterministic component and the visual-evaluation contract. Document and spreadsheet
tasks follow the same packaging protocol and are scored exclusively by
task-specific deterministic judges. Replays of the reference solutions check
execution, while model--harness evaluations measure agent performance
(Table~\ref{tab:static-main-results}).

Using criteria specific to each task is also studied in agent-trajectory evaluation
\citep{ding2026adarubrictaskadaptiverubricsreliable}. Their specificity alone does
not establish judge reliability: rubric verifiers can still misread complex agent
outputs \citep{peng2026llmasajudgereliablyverifyrubrics}. We therefore distinguish
implementation checks from human agreement in
Section~\ref{sec:judge-reliability} and Appendix~\ref{app:judge-reliability},
where both are reported per modality.

\begin{figure}[!t]
    \centering
    \includegraphics[width=\linewidth]{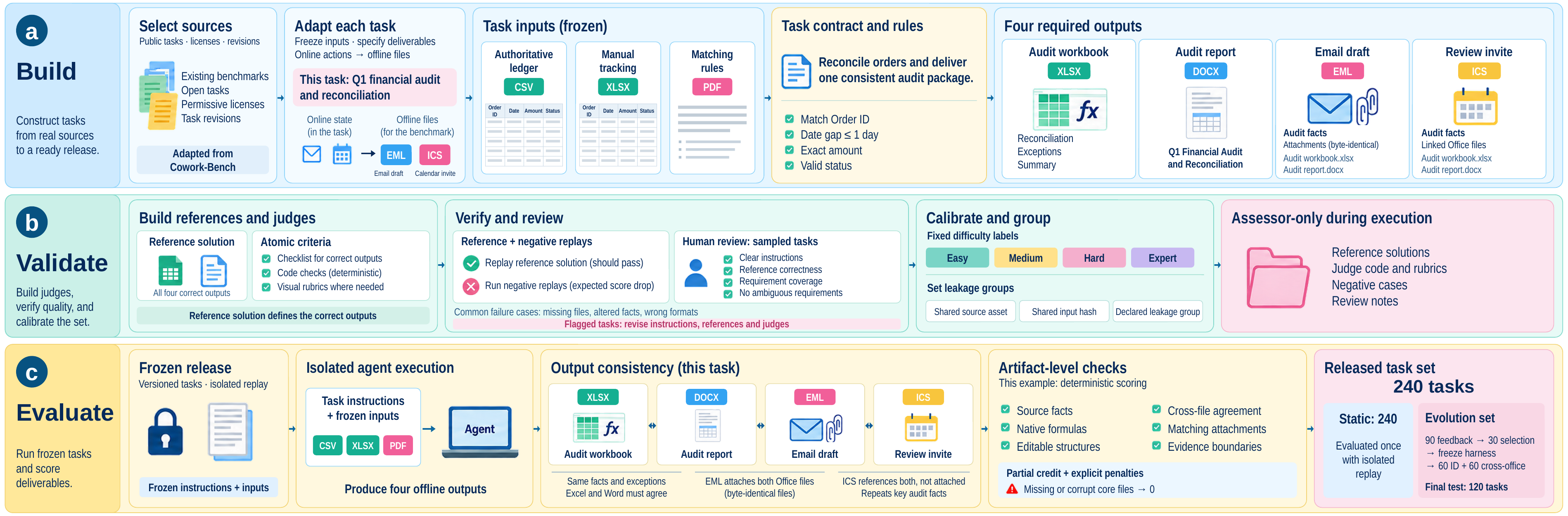}
    \caption{A cross-office audit adapted from Cowork-Bench illustrates
    (a) task construction, (b) validation of the reference solution and rubric,
    and (c) artifact evaluation for correctness, editability, and cross-file
    consistency across the four deliverables.}
    \label{fig:cross-office-task-example}
\end{figure}

\subsection{Replay Controls and Data Access}

Static evaluation keeps the model--harness configuration, task inputs, judges,
attempt policy, and resource limits fixed for the corresponding comparison.
The agent receives the instruction and allowed inputs, while reference answers
and verifier internals remain outside its writable workspace. Saved outputs,
execution traces, and task-level scores support later auditing. The primary static
diagnostic covers all 240 tasks. Evolution uses separate feedback, selection,
and final partitions (Appendix~\ref{app:split-boundaries}). Harness changes are
made during feedback and selected on validation tasks; final evaluation freezes
the selected revision. Static and final evolution scores are reported on
their respective task sets.

\section{Results and Analysis}

\subsection{Experimental Setup}
\label{sec:experimental-setup}

\paragraph{Static evaluation.}
(a)~\textbf{Benchmark and baselines.}
The static evaluation covers all 240 tasks. Table~\ref{tab:static-main-results}
compares Sonnet~5 and GLM~5.3 with Claude Code, OpenCode, OpenHands, and
mini-SWE-agent (MSA). GLM~5.3 is excluded from difficulty
calibration and uses the fixed Sonnet~5 labels. Versions, score reuse, and
reporting details appear in Appendix~\ref{app:static-evaluation-details}.
(b)~\textbf{Metrics.}
Scores are 100 times mean task reward: each modality averages its 60 tasks,
and the overall score averages the four modality means.
Valid agent failures receive the task judge's score; unresolved
infrastructure failures are recorded separately in the run records.

\paragraph{Evolution evaluation.}
\label{sec:evolution-setup}
(a)~\textbf{Baselines.}
The comparison evaluates ACE~\citep{zhang2025ace} and
AHE~\citep{lin2026ahe} against a frozen MSA harness $H_0$, with GLM~5.3
as the solver and updater. ACE updates a context playbook; AHE revises harness
components using execution evidence. Both start from the same harness.
(b)~\textbf{Protocol.}
The protocol separates feedback, revision selection, and final testing. Each
selected harness is frozen before evaluation on held-out single-modality and
cross-office tasks, using the same task versions, inference budgets, and judges
as $H_0$. Appendix~\ref{app:split-boundaries} gives the split and access
rules; Appendix~\ref{app:evolution-evaluation-details} describes the shared budget
controls and configuration requirements.

\subsection{Static Model--Harness Comparison}

Table~\ref{tab:static-main-results} reports the completed model--harness results.
With Sonnet~5, MSA scores 38.08 overall, OpenHands 36.73, Claude Code 33.85,
and OpenCode 32.47. Within that panel, each harness scores lower on cross-office tasks than on
any single modality. These are mean rewards, not percentages of tasks solved;
spreadsheet judges can also assign negative penalties. The 240-task Claude Code run with GLM~5.3
scores 42.25 overall and 30.62 on cross-office tasks. MSA scores 39.40 overall and 28.68
on cross-office tasks; OpenCode and OpenHands score 25.04 and 22.21 overall,
respectively. All GLM~5.3 results use the fixed difficulty labels.

Figure~\ref{fig:expert-model-comparison} compares five models on the 48 Expert
tasks with a frozen MSA harness. GPT6 Astra leads in presentations and Kimi K3
in documents and spreadsheets, while all five models score 3.54--6.06 on
cross-office tasks under the same protocol.

\begin{table}[t]
    \centering
    \begingroup
    \includegraphics[width=\linewidth]{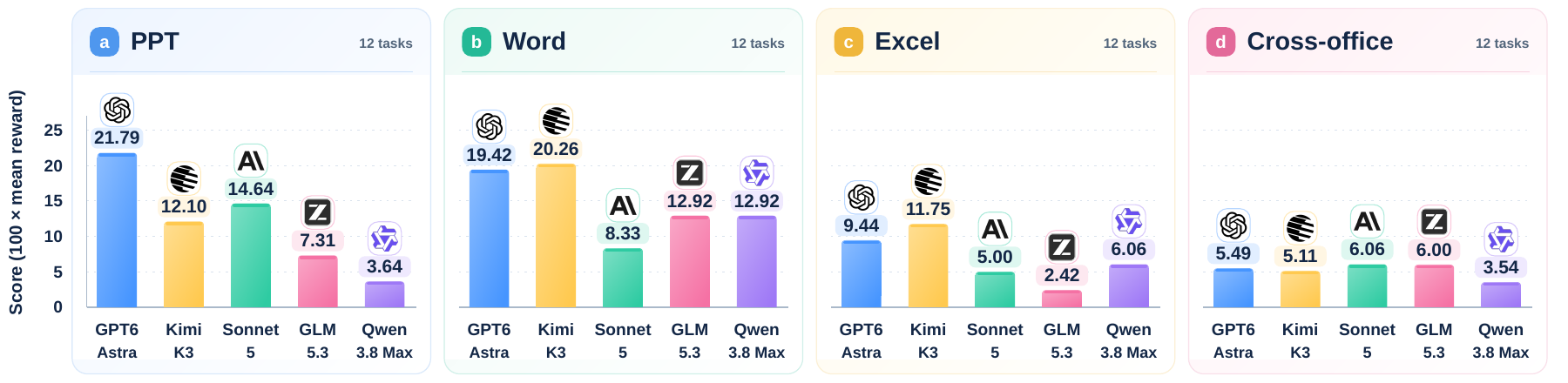}
    \captionof{figure}{Expert-tier performance with a frozen MSA harness. Each panel
    reports 100 $\times$ mean reward over the same 12 tasks for all five models.}
    \label{fig:expert-model-comparison}
    \endgroup
    \par\vspace{8pt}
    \caption{Static performance by modality and difficulty (100 $\times$ mean reward).
    All summarizes 60 tasks per modality; tiers follow the calibration described in
    Appendix~\ref{app:difficulty-taxonomies}. Overall averages the four modality
    means at each difficulty level. MSA denotes mini-SWE-agent. Each reported
    configuration covers all 240 tasks.}
    \label{tab:static-main-results}
    \footnotesize
    \setlength{\tabcolsep}{2pt}
    \renewcommand{\arraystretch}{1.10}
    \begin{tabular*}{\linewidth}{@{\extracolsep{\fill}}ll>{\columncolor{tablegray}}rrrrr@{\hspace{6pt}}>{\columncolor{tablegray}}rrrrr@{}}
        \toprule
        & & \multicolumn{5}{c}{\textbf{Sonnet 5}} & \multicolumn{5}{c}{\textbf{GLM 5.3}} \\
        \cmidrule(lr){3-7}\cmidrule(lr){8-12}
        \multirow{-2}{*}{Harness} & \multirow{-2}{*}{Modality}
        & \textbf{All} & Easy & Medium & Hard & Expert
        & \textbf{All} & Easy & Medium & Hard & Expert \\
        \midrule
         & PPT & \textbf{37.29} & 96.96 & 35.51 & 18.02 & 9.19 & \textbf{49.16} & 99.29 & 64.97 & 25.77 & 10.38 \\
         & Word & \textbf{36.59} & 90.59 & 50.21 & 11.37 & 0.00 & \textbf{42.48} & 80.46 & 62.75 & 19.00 & 9.33 \\
         & Excel & \textbf{38.38} & 79.45 & 51.33 & 23.58 & 0.09 & \textbf{46.76} & 81.08 & 60.75 & 29.81 & 16.89 \\
         & Cross-office & \textbf{23.12} & 72.20 & 18.61 & 10.32 & 0.02 & \textbf{30.62} & 72.58 & 27.09 & 22.83 & 5.63 \\
        \multirow{-5}{*}{Claude Code} & \textbf{Overall} & \textbf{33.85} & \textbf{84.80} & \textbf{38.91} & \textbf{15.82} & \textbf{2.32} & \textbf{42.25} & \textbf{83.35} & \textbf{53.89} & \textbf{24.35} & \textbf{10.56} \\
        \midrule
         & PPT & \textbf{36.87} & 97.29 & 37.27 & 16.53 & 6.37 & \textbf{40.90} & 93.38 & 59.02 & 12.09 & 4.47 \\
         & Word & \textbf{38.62} & 84.13 & 59.52 & 13.14 & 0.00 & \textbf{36.61} & 73.69 & 59.00 & 8.36 & 8.33 \\
         & Excel & \textbf{34.03} & 69.25 & 36.65 & 26.31 & 6.47 & \textbf{15.76} & 33.12 & 21.78 & 11.49 & -4.25 \\
         & Cross-office & \textbf{20.34} & 71.91 & 15.93 & 3.93 & 0.00 & \textbf{6.88} & 21.66 & 4.35 & 4.13 & 0.00 \\
        \multirow{-5}{*}{OpenCode} & \textbf{Overall} & \textbf{32.47} & \textbf{80.64} & \textbf{37.34} & \textbf{14.98} & \textbf{3.21} & \textbf{25.04} & \textbf{55.47} & \textbf{36.04} & \textbf{9.02} & \textbf{2.14} \\
        \midrule
         & PPT & \textbf{38.07} & 97.98 & 40.61 & 16.23 & 7.11 & \textbf{20.71} & 68.72 & 12.70 & 7.79 & 4.11 \\
         & Word & \textbf{40.60} & 85.99 & 66.77 & 11.22 & 0.00 & \textbf{37.30} & 68.77 & 53.94 & 10.39 & 21.25 \\
         & Excel & \textbf{44.98} & 83.29 & 61.61 & 27.44 & 8.00 & \textbf{12.16} & 42.65 & 10.47 & 7.65 & -9.06 \\
         & Cross-office & \textbf{23.26} & 65.51 & 24.69 & 8.99 & 0.26 & \textbf{18.66} & 48.18 & 16.48 & 10.84 & 4.12 \\
        \multirow{-5}{*}{OpenHands} & \textbf{Overall} & \textbf{36.73} & \textbf{83.20} & \textbf{48.42} & \textbf{15.97} & \textbf{3.84} & \textbf{22.21} & \textbf{57.08} & \textbf{23.40} & \textbf{9.17} & \textbf{5.11} \\
        \midrule
         & PPT & \textbf{47.92} & 98.19 & 62.93 & 21.59 & 14.64 & \textbf{47.58} & 99.07 & 75.94 & 11.73 & 7.31 \\
         & Word & \textbf{38.13} & 84.10 & 53.72 & 11.75 & 8.33 & \textbf{41.06} & 78.08 & 60.46 & 15.73 & 12.92 \\
         & Excel & \textbf{38.30} & 64.57 & 59.88 & 21.39 & 5.00 & \textbf{40.29} & 73.42 & 59.91 & 23.81 & 2.42 \\
         & Cross-office & \textbf{27.99} & 74.75 & 24.62 & 14.80 & 6.06 & \textbf{28.68} & 69.29 & 26.70 & 18.69 & 6.00 \\
        \multirow{-5}{*}{MSA} & \textbf{Overall} & \textbf{38.08} & \textbf{80.40} & \textbf{50.29} & \textbf{17.38} & \textbf{8.51} & \textbf{39.40} & \textbf{79.97} & \textbf{55.76} & \textbf{17.49} & \textbf{7.16} \\
        \bottomrule
    \end{tabular*}
\end{table}

\subsection{Evolution and Cross-Office Transfer}

Table~\ref{tab:evolution-main-results} compares the frozen MSA harness $H_0$
with the ACE and AHE revisions on 120 final tasks: 60 in-domain tasks (20 each
for presentations, documents, and spreadsheets) and 60 cross-office tasks. All
three use GLM~5.3 with matched per-task limits and scoring rules. ACE scores
39.72 overall (+2.46 over $H_0$) and AHE 37.72 (+0.46), with in-domain gains of
4.90 and 2.64, respectively. Cross-office scores are 28.70 for ACE (+0.03) and
26.96 for AHE ($-1.72$). AHE crashes in 31/120 final attempts, including 17
cross-office cases, in its added completion check; the crash-excluded diagnostic
appears in Table~\ref{tab:ahe-crash-split}. Both methods therefore raise overall
performance while cross-office scores remain nearly unchanged or decline,
which the separate test sets make visible.

Selection scores also differ from final gains: ACE improves by 7.55 points on
the 30 selection tasks, whereas AHE loses 7.03 yet improves slightly on final
tasks. These outcomes support separate selection and final tests and direct
measurement of cross-office transfer. We report AHE's evolved revision even
though its original selection rule retained $H_0$; the reporting rule is
disclosed in Appendix~\ref{app:evolution-evaluation-details}. Each method
contributes one fixed-model run.

To calibrate run-to-run variation, we rerun frozen $H_0$ on the 60 cross-office
final tasks with a different sampling seed. The rerun scores 28.31, 0.37 points
below 28.68, our noise scale for one run. ACE's $+0.03$ lies within this noise;
AHE's $-1.72$ is roughly five times larger.

\begin{table}[t]
    \centering
    \caption{Evolution results (GLM~5.3, MSA seed). Scores are 100$\times$ mean
    reward; $\Delta$ is the change from $H_0$ on the same tasks. Selection and
    final sets are disjoint. AHE reports the evolved revision, without fallback
    to $H_0$.}
    \label{tab:evolution-main-results}
    \small
    \setlength{\tabcolsep}{3pt}
    \begin{tabular*}{\linewidth}{@{\extracolsep{\fill}}lrrrrrrrr@{}}
        \toprule
        & \multicolumn{2}{c}{Selection (30)}
        & \multicolumn{2}{c}{In-domain (60)}
        & \multicolumn{2}{c}{Cross-office (60)}
        & \multicolumn{2}{c}{All (120)} \\
        \cmidrule(lr){2-3}\cmidrule(lr){4-5}\cmidrule(lr){6-7}\cmidrule(lr){8-9}
        Method & Score & $\Delta$ & Score & $\Delta$ & Score & $\Delta$ & Score & $\Delta$ \\
        \midrule
        \graytablerow
        Frozen $H_0$ & 47.44 & 0.00 & 45.83 & 0.00 & 28.68 & 0.00 & 37.25 & 0.00 \\
        ACE~\citep{zhang2025ace} & 54.98 & \gainpoints{7.55} & 50.73 & \gainpoints{4.90} & 28.70 & \gainpoints{0.03} & 39.72 & \gainpoints{2.46} \\
        AHE~\citep{lin2026ahe} & 40.41 & \losspoints{7.03} & 48.47 & \gainpoints{2.64} & 26.96 & \losspoints{1.72} & 37.72 & \gainpoints{0.46} \\
        \bottomrule
    \end{tabular*}
\end{table}

\subsection{Judge Reliability and Human Audit}
\label{sec:judge-reliability}

Table~\ref{tab:judge-reliability} reports checks on the full 240-task release
alongside the human audit. All reference solutions receive a deterministic
reward of 1. Empty and corrupt outputs receive exactly 0 on all presentation and
cross-office tasks, but some document and spreadsheet judges retain positive
credit without a valid deliverable. Spreadsheet penalties also produce negative
rewards; these count as deviations from the expected zero score in this table.
Appendix~\ref{app:judge-reliability} separates these outcomes and describes the
checks and human audit protocol.

The human audit rates 60 outputs per modality with three raters each. Ordinal
Krippendorff's $\alpha$ ranges from 0.969 to 0.994, so the panels apply the
rubrics consistently and the adjudicated consensus rests on a stable target.
Score MAE against that consensus is 3.779 points on spreadsheets, 7.393 on
cross-office workflows, 8.501 on documents, and 9.576 on presentations, on the
100-times-reward scale. The ordering runs opposite to the implementation checks:
spreadsheet judges retain the most credit for invalid outputs yet track human
scores most closely, while presentation judges pass every deterministic check and
still differ most from human readers. Deterministic checks and human agreement
therefore measure different things, and neither substitutes for the other.

\begin{table}[t]
    \begingroup
    \centering
    \caption{Deterministic judge checks and human audit. Implementation checks
    report the fraction of the 60 tasks that receive the expected reward; values
    below 1.00 cover tasks scoring above or below it.}
    \label{tab:judge-reliability}
    \small
    \setlength{\tabcolsep}{3pt}
    \renewcommand{\arraystretch}{1.05}
    \begin{tabular*}{\linewidth}{@{\extracolsep{\fill}}lr*{3}{>{\columncolor{tablegray}}c}cccc@{}}
        \toprule
        & & \multicolumn{3}{c}{\cellcolor{tablegray}Implementation checks} & \multicolumn{4}{c}{Human audit} \\
        \cmidrule(lr){3-5}\cmidrule(lr){6-9}
        Modality & Tasks
        & \shortstack{Ref.\ replay\\($r=1$)}
        & \shortstack{Empty out\\($r=0$)}
        & \shortstack{Corrupt out\\($r=0$)}
        & Outputs & Raters & IAA $\alpha$
        & \shortstack{Score\\MAE} \\
        \midrule
        Presentation & 60 & 1.00 & 1.00 & 1.00 & 60 & 3 & 0.969 & 9.576 \\
        Document     & 60 & 1.00 & 0.88 & 0.88 & 60 & 3 & 0.994 & 8.501 \\
        Spreadsheet  & 60 & 1.00 & 0.45 & 0.43 & 60 & 3 & 0.981 & 3.779 \\
        Cross-office & 60 & 1.00 & 1.00 & 1.00 & 60 & 3 & 0.983 & 7.393 \\
        \bottomrule
    \end{tabular*}
    \endgroup
    \par\vspace{8pt}
    
\begingroup
    \centering
    \caption{Delivery failures and artifact-quality violations. The upper panel
    covers static configurations; the lower panel compares evolution methods on
    the same 120 final tasks. Entries are proportions on a 0--1 scale, rounded to
    three decimal places. Quality columns count rule-identified violations over
    240 tasks per static configuration or 120 tasks per evolution configuration.
    Human reviewers confirmed the category assignments on a stratified sample\linebreak (Appendix~\ref{app:failure-classification-review}). A task can fall in several
    categories, so the columns do not sum.}
    \label{tab:failure-analysis}
    \begingroup
    \color{black}
    \definecolor{failureband}{gray}{0.94}
    \newcommand{\fhead}[1]{\textbf{#1}}
    \newcommand{\fcolhead}[1]{{\fontsize{6.5}{8}\selectfont\fhead{#1}}}
    \newcommand{\fquality}{\fpending & \fpending & \fpending & \fpending &
        \fpending & \fpending & \fpending}
    \newcommand{\ftotal}[1]{\cellcolor{failureband}\textbf{#1}}
    \sffamily
    \fontsize{7}{9}\selectfont
    \setlength{\tabcolsep}{1.15pt}
    \renewcommand{\arraystretch}{1.36}
    \arrayrulecolor{black}
    \begin{tabular*}{\linewidth}{@{\extracolsep{\fill}}l@{\hspace{4pt}}lccccc!{\vrule width 0.35pt}ccccccc@{}}
        \toprule[0.8pt]
        & & \multicolumn{5}{c}{\fhead{Delivery failures $\downarrow$}}
        & \multicolumn{7}{c}{\fhead{Artifact-quality violations $\downarrow$}} \\
        \cmidrule(lr){3-7}\cmidrule(lr){8-14}
        \multirow{-2}{*}{\fhead{Model}}
        & \multirow{-2}{*}{\fhead{Harness}}
        & \fhead{PPT} & \fhead{Word} & \fhead{Excel}
        & \fhead{\shortstack{Cross-\\office}} & \ftotal{All}
        & \fcolhead{\shortstack{Facts /\\calcs}}
        & \fcolhead{\shortstack{Native\\structure}}
        & \fcolhead{\shortstack{Protected\\content}}
        & \fcolhead{\shortstack{Evidence\\trace}}
        & \fcolhead{\shortstack{Scope /\\privacy}}
        & \fcolhead{\shortstack{Cross-file\\consist.}}
        & \fcolhead{\shortstack{Visual\\layout}} \\
        \midrule
        \rowcolor{failureband}
        \multicolumn{14}{@{}l@{}}{\rule{0pt}{14pt}\hspace{4pt}
            \fhead{STATIC EVALUATION}\quad
            Frozen model--harness configurations\enspace$\cdot$\enspace240 tasks each} \\
        \midrule
        & Claude Code & 0.000 & 0.250 & 0.117 & 0.100
        & \ftotal{0.117} & 0.517 & 0.204 & 0.042 & 0.037 & 0.075 & 0.021 & 0.154 \\
        & OpenCode & 0.017 & 0.217 & 0.183 & 0.267
        & \ftotal{0.171} & 0.442 & 0.200 & 0.058 & 0.029 & 0.058 & 0.017 & 0.142 \\
        & OpenHands & 0.000 & 0.233 & 0.067 & 0.117
        & \ftotal{0.104} & 0.483 & 0.221 & 0.054 & 0.029 & 0.071 & 0.021 & 0.163 \\
        \multirow{-4}{*}{\fhead{Sonnet 5}}
        & MSA & 0.200 & 0.350 & 0.217 & 0.200
        & \ftotal{0.242} & 0.492 & 0.129 & 0.021 & 0.063 & 0.083 & 0.013 & 0.075 \\
        \cmidrule(lr){1-14}
        & Claude Code & 0.150 & 0.167 & 0.100 & 0.050
        & \ftotal{0.117} & 0.558 & 0.154 & 0.029 & 0.079 & 0.108 & 0.025 & 0.096 \\
        & OpenCode & 0.450 & 0.424 & 0.467 & 0.867
        & \ftotal{0.552} & 0.154 & 0.050 & 0.004 & 0.008 & 0.008 & 0.000 & 0.033 \\
        & OpenHands & 0.650 & 0.417 & 0.667 & 0.517
        & \ftotal{0.563} & 0.233 & 0.071 & 0.004 & 0.042 & 0.063 & 0.008 & 0.033 \\
        \multirow{-4}{*}{\fhead{GLM 5.3}}
        & MSA & 0.283 & 0.217 & 0.233 & 0.267
        & \ftotal{0.250} & 0.442 & 0.121 & 0.017 & 0.054 & 0.087 & 0.017 & 0.054 \\
        \midrule
        \rowcolor{failureband}
        \multicolumn{14}{@{}l@{}}{\rule{0pt}{14pt}\hspace{4pt}
            \fhead{EVOLUTION EVALUATION}\quad
            Shared MSA seed\enspace$\cdot$\enspace Same 120 final tasks} \\
        \midrule
        & MSA ($H_0$) & 0.250 & 0.300 & 0.250 & 0.267
        & \ftotal{0.267} & 0.508 & 0.108 & 0.008 & 0.092 & 0.158 & 0.033 & 0.058 \\
        & MSA + ACE & 0.150 & 0.250 & 0.200 & 0.233
        & \ftotal{0.217} & 0.542 & 0.125 & 0.042 & 0.092 & 0.150 & 0.033 & 0.042 \\
        \multirow{-3}{*}{\fhead{GLM 5.3}}
        & MSA + AHE & 0.050 & 0.000 & 0.200 & 0.283
        & \ftotal{0.183} & 0.558 & 0.158 & 0.033 & 0.083 & 0.133 & 0.025 & 0.100 \\
        \bottomrule[0.8pt]
    \end{tabular*}
    \arrayrulecolor{black}
    \endgroup
\endgroup

\end{table}

\subsection{Failure Types}

Table~\ref{tab:failure-analysis} separates static configurations from evolved
harnesses on the same final tasks. Delivery columns count omitted required files;
quality columns classify judge-recorded requirements into seven rule-based
categories, with an agent-assisted pass where no rule matches. Human reviewers
confirmed the assignments on 160 stratified cases (Appendix~\ref{app:failure-classification-review}),
with no errors. Missing-file checks are not counted as additional violations;
coverage and denominators are in Appendix~\ref{app:failure-cost}.

Incorrect facts and calculations are the largest category in all eleven
configurations, covering 0.154 to 0.558 of tasks and exceeding the
delivery-failure rate in nine of them. Invalid native structure is second in
nine configurations, at 0.050 to 0.221; for $H_0$ and ACE, unrequested scope or
privacy changes take second place. Unrequested changes to protected content and
cross-file inconsistency stay at or below 0.058 everywhere. The evolution panel
separates delivery from content: ACE and AHE both deliver more files than $H_0$,
lowering the delivery-failure rate from 0.267 to 0.217 and 0.183, while
violations rise in the content categories --- incorrect facts and calculations go
from 0.508 to 0.542 for ACE and 0.558 for AHE, invalid native structure from
0.108 to 0.125 and 0.158, and protected-content changes from 0.008 to 0.042 and
0.033 --- partly because more delivered files bring more checks into scope, while
scope and cross-file violations stay flat or fall.

\subsection{Difficulty Generalization to a Held-Out Model}

Across four GLM~5.3 harnesses and 240 tasks, tier means remain non-increasing
under fixed Sonnet~5 labels, except at the document Hard--Expert boundary:
all 12 Expert tasks have zero calibration scores and require tie-breaking. We keep
these labels fixed for GLM~5.3 to test transfer across models without rebuilding
the tiers. The exception matches its cause: the document Expert tier was cut
from Hard by tie-breaking within a block of zero calibration scores, and under
GLM~5.3 one harness scores higher on Expert than on Hard documents, while every
boundary backed by measured scores stays ordered. Expert tasks also remain
hardest for GLM~5.3 itself, at 2.14--10.56 across its four harnesses, so the
tiers track difficulty for both models rather than the calibration panel alone.

\section*{AI Use Statement}

AI tools assisted with manuscript proofreading, language editing, reference
verification, and dataset construction, including task synthesis and adaptation.
The authors are responsible for verifying the resulting text, citations, dataset
content, and reported experimental results.

\section*{Ethics Statement}

Task packages should preserve upstream attribution and license obligations.
Adapted business scenarios and automated scores are research evaluation material,
not authorization to act on real accounts or a substitute for professional review.
Any release of artifacts or traces must exclude credentials and review source
materials for personal or otherwise restricted information.

\bibliography{iclr2027_conference}
\bibliographystyle{iclr2027_conference}

\clearpage
\appendix
\etocdepthtag.toc{appendix}

\begingroup

\definecolor{appendixtoclink}{HTML}{617A70}
\hypersetup{linkcolor=appendixtoclink}
\linespread{1.45}\selectfont
\setlength{\parskip}{0pt}
\makeatletter
\renewcommand*\l@section{\addvspace{0.5em}\bfseries\@dottedtocline{1}{0em}{1.5em}}
\renewcommand*\l@subsection{\@dottedtocline{2}{1.5em}{2.4em}}
\makeatother
\etocsettagdepth{main}{none}
\etocsettagdepth{appendix}{subsection}
\etocsettocstyle{
    \section*{Contents}
    \pdfbookmark[0]{Appendix Contents}{appendix-contents}
}{}
\tableofcontents
\endgroup
\hypersetup{linkcolor=blue}
\clearpage
\setcounter{table}{0}
\renewcommand{\thetable}{A\arabic{table}}
\renewcommand{\theHtable}{A\arabic{table}}
\setcounter{figure}{0}
\renewcommand{\thefigure}{A\arabic{figure}}
\renewcommand{\theHfigure}{A\arabic{figure}}

\section{Structural Regimes and Difficulty Calibration}
\label{app:difficulty-taxonomies}

Each task is described by its workflow category and empirical difficulty.
Workflow categories are modality-specific; difficulty uses the same calibration
protocol across all four modalities.

\subsection{Common Empirical Difficulty Calibration}
\label{app:difficulty-calibration}

Empirical calibration uses one frozen panel of model--harness configurations,
keeping model versions, task packages, judges, seed policy, evaluation budget, and
resource accounting fixed. Infrastructure failures are retried or excluded; a valid
run that omits a required deliverable receives the score the judge assigns. Tasks
are sorted by $\bar r_i$ and partitioned into precommitted 20/30/30/20 rank bins.
A tie at a boundary is resolved first by structural regime and then by stable task
ID. For tier $\ell$ with task set $\mathcal{T}_{\ell}$, the calibration statistic is
\begin{equation}
    \bar r_{\ell}
    =\frac{1}{|\mathcal{T}_{\ell}|}
      \sum_{i\in\mathcal{T}_{\ell}}\bar r_i.
    \label{eq:tier-calibration}
\end{equation}
The workflow regime describes task requirements and serves as a tie-breaking
reference; it is not an alternative difficulty scale.

The calibration panel uses Claude Code, OpenCode, and OpenHands with the
Sonnet~5 backbone. Each task contributes one selected completion per harness, and
within-modality ranks define the difficulty labels reported here.

\subsection{Presentation}

Presentation structure is annotated from five observable properties: operation scope,
evidence dependency, reasoning and synthesis, PowerPoint structural depth, and
constraint coupling. Slide count alone is not a difficulty criterion. These properties
place a task into one of four workflow regimes:

\begin{figure}[H]
    \centering
    \includegraphics[width=\linewidth]{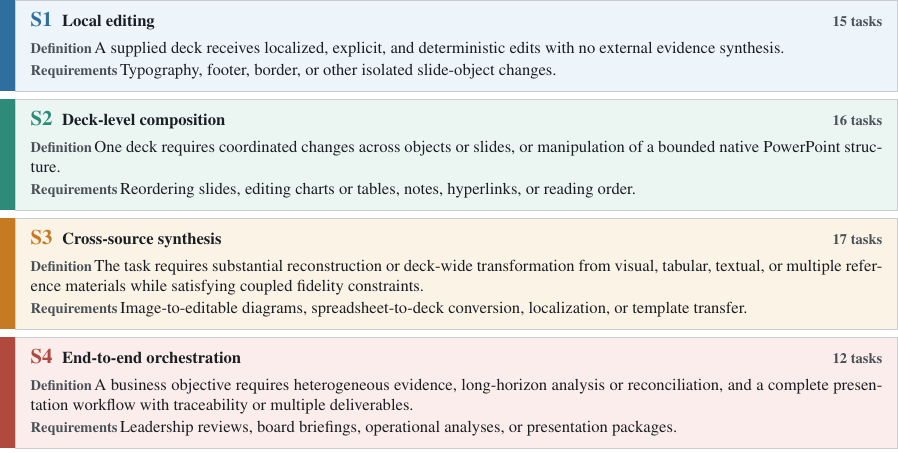}
    \caption{Structural workflow regimes for presentation tasks, with the current
    structural audit counts. S1--S4 are not the empirical difficulty tiers.}
    \label{fig:presentation-regimes}
\end{figure}

The per-regime counts in Figure~\ref{fig:presentation-regimes} measure workflow
coverage. Table~\ref{tab:presentation-calibration} reports the resulting rank partition.

\begin{table}[H]
    \centering
    \caption{Hybrid-reward calibration of the 60 presentation tasks. Values are
    percentages; the four rows are precommitted 20/30/30/20 rank bins.}
    \label{tab:presentation-calibration}
    \scriptsize
    \resizebox{\linewidth}{!}{
    \begin{tabular}{lrrrrrr}
        \toprule
        Tier & Tasks & Claude Code & OpenCode & OpenHands & Panel mean $\bar r_{\ell}$ & $1-\bar r_{\ell}$ \\
        \midrule
        Easy   & 12 & 96.96 & 97.29 & 97.98 & 97.41 &  2.59 \\
        Medium & 18 & 35.51 & 37.27 & 40.61 & 37.80 & 62.20 \\
        Hard   & 18 & 18.02 & 16.53 & 16.23 & 16.93 & 83.07 \\
        Expert & 12 &  9.19 &  6.37 &  7.11 &  7.56 & 92.44 \\
        \bottomrule
    \end{tabular}}
\end{table}

Deterministic judges are combined with task-specific visual rubrics. Of 180 scoring
attempts, 179 contained the required deliverable; OpenCode omitted the required PPTX
for one task. The panel mean is 37.41\%.

\subsection{Document}

\begin{figure}[H]
    \centering
    \includegraphics[width=\linewidth]{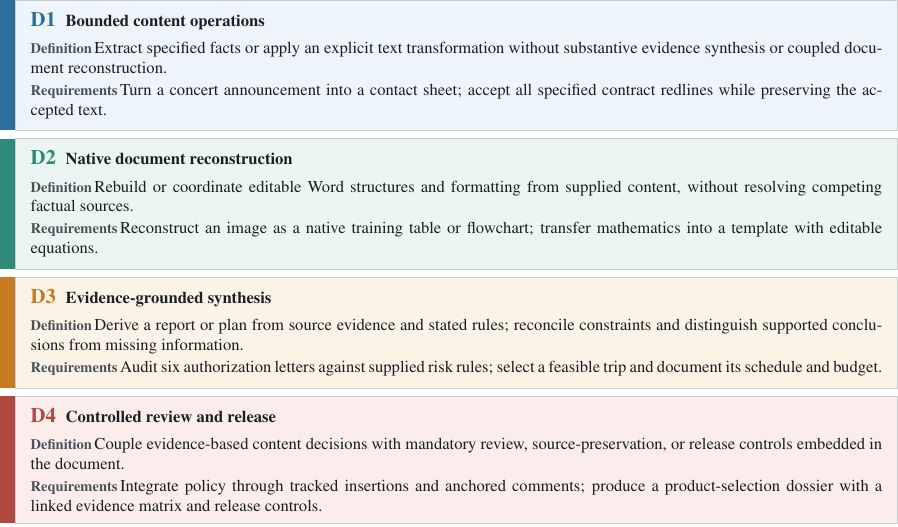}
    \caption{Structural workflow regimes for document tasks, with examples from the
    current pool. D1--D4 are not the empirical difficulty tiers.}
    \label{fig:document-regimes}
\end{figure}

Document structure is characterized by the scope of content changes, the depth of
native Word objects, evidence synthesis, preservation constraints, and review or
release controls. Figure~\ref{fig:document-regimes} distinguishes four workflow
regimes using requirements from the current task pool. Page count and the number of
input files do not determine the regime.

The required workflow determines the regime: native-object complexity alone belongs
to D2, substantive evidence synthesis to D3, and a mandatory controlled-review or
release process to D4. A comment or bookmark alone does not imply D4. Inputs may
include PDFs, images, spreadsheets, or emails while the required deliverable
remains a Word document; heterogeneous inputs alone do not make a task cross-office.
Table~\ref{tab:document-calibration} reports the empirical partition.

\begin{table}[H]
    \centering
    \caption{Deterministic-reward pilot calibration of the 60 document tasks.
    Values are percentages; the four rows use 20/30/30/20 rank bins.}
    \label{tab:document-calibration}
    \scriptsize
    \resizebox{\linewidth}{!}{
    \begin{tabular}{lrrrrrr}
        \toprule
        Tier & Tasks & Claude Code & OpenCode & OpenHands & Panel mean $\bar r_{\ell}$ & $1-\bar r_{\ell}$ \\
        \midrule
        Easy   & 12 & 90.59 & 84.13 & 85.99 & 86.90 &  13.10 \\
        Medium & 18 & 50.21 & 59.52 & 66.77 & 58.83 &  41.17 \\
        Hard   & 18 & 11.37 & 13.14 & 11.22 & 11.91 &  88.09 \\
        Expert & 12 &  0.00 &  0.00 &  0.00 &  0.00 & 100.00 \\
        \bottomrule
    \end{tabular}}
\end{table}

The same Sonnet~5 panel contributes 17 fresh upgraded-task evaluations and 43 reused,
reused historical scores for unchanged tasks per harness; document scoring is deterministic
only. The panel mean is 38.60\%. Twenty tasks receive zero across all three
harnesses, including all 12 Expert tasks.

\subsection{Spreadsheet}

\begin{figure}[H]
    \centering
    \includegraphics[width=\linewidth]{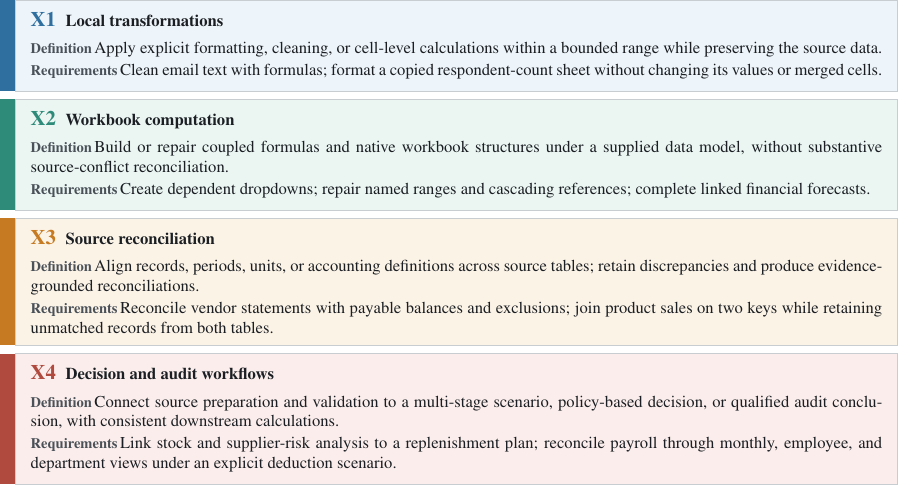}
    \caption{Structural workflow regimes for spreadsheet tasks, with examples from
    the current pool. X1--X4 are not the empirical difficulty tiers.}
    \label{fig:spreadsheet-regimes}
\end{figure}

Spreadsheet structure is characterized by the scope of cell operations, formula and
cross-sheet dependencies, alignment of source entities and periods, analytical
reasoning, and the coupling of assumptions to downstream decisions.
Figure~\ref{fig:spreadsheet-regimes} separates four workflow regimes. A single-file
model can require deep dependency repair, whereas a large table can require only a
local transformation; file, sheet, and row counts are therefore not sufficient
classification rules.

The category depends on the required sequence of operations: a supplied financial
model repair can be X2, an evidence reconciliation X3, and a workflow that additionally
couples scenarios or operational decisions to those checks X4. A chart or summary
sheet alone does not imply X4. Workbook-internal email drafts or recommendations do
not create an additional Office modality. Spreadsheet scores use deterministic
judges, and Easy--Expert membership follows the same three-harness calibration
protocol used for the other modalities.

\begin{table}[H]
    \centering
    \caption{Deterministic-reward pilot calibration of the 60 spreadsheet tasks.
    Rewards and empirical difficulty are scaled by 100; the four rows use
    20/30/30/20 rank bins.}
    \label{tab:spreadsheet-calibration}
    \scriptsize
    \resizebox{\linewidth}{!}{
    \begin{tabular}{lrrrrrr}
        \toprule
        Tier & Tasks & Claude Code & OpenCode & OpenHands & Panel mean $\bar r_{\ell}$ & $1-\bar r_{\ell}$ \\
        \midrule
        Easy   & 12 & 79.45 & 69.25 & 83.29 & 77.33 & 22.67 \\
        Medium & 18 & 51.33 & 36.65 & 61.61 & 49.86 & 50.14 \\
        Hard   & 18 & 23.58 & 26.31 & 27.44 & 25.78 & 74.22 \\
        Expert & 12 &  0.09 &  6.47 &  8.00 &  4.85 & 95.15 \\
        \midrule
        All    & 60 & 38.38 & 34.03 & 44.98 & 39.13 & 60.87 \\
        \bottomrule
    \end{tabular}}
\end{table}

Table~\ref{tab:spreadsheet-calibration} uses 180 valid scores: per harness, seven
upgraded tasks are newly executed and 53 unchanged tasks reuse saved outputs
with task and verifier versions matched. The panel mean is 39.13\%.

\subsection{Cross-Office}

Cross-office structure is annotated from five observable properties: the number and
heterogeneity of applications and artifacts, dependency-graph depth, evidence
reconciliation, coupling among deliverables, and workflow and review requirements.
These properties place a task into one of four workflow regimes:

\begin{figure}[H]
    \centering
    \includegraphics[width=\linewidth]{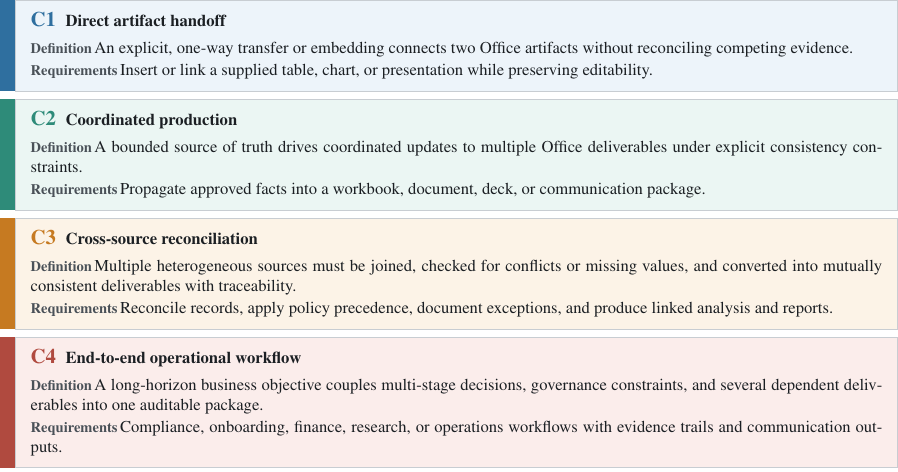}
    \caption{Structural workflow regimes for cross-office tasks. C1--C4 are not the
    empirical difficulty tiers.}
    \label{fig:crossoffice-regimes}
\end{figure}

The four regimes explain the observable structure of a task; the final ordinal label
is assigned by the shared empirical protocol. Table~\ref{tab:cross-office-calibration}
reports the current 20/30/30/20 rank partition.

\begin{table}[H]
    \centering
    \caption{Hybrid-reward calibration of the 60 cross-office tasks.}
    \label{tab:cross-office-calibration}
    \scriptsize
    \resizebox{\linewidth}{!}{
    \begin{tabular}{lrrrrrr}
        \toprule
        Tier & Tasks & Claude Code & OpenCode & OpenHands & Panel mean $\bar r_{\ell}$ & $1-\bar r_{\ell}$ \\
        \midrule
        Easy   & 12 & 72.20 & 71.91 & 65.51 & 69.87 & 30.13 \\
        Medium & 18 & 18.61 & 15.93 & 24.69 & 19.74 & 80.26 \\
        Hard   & 18 & 10.32 &  3.93 &  8.99 &  7.75 & 92.25 \\
        Expert & 12 &  0.02 &  0.00 &  0.26 &  0.09 & 99.91 \\
        \bottomrule
    \end{tabular}}
\end{table}

Thirty-six tasks use the 70/30 deterministic--visual reward and 24 use deterministic
reward only. Of 180 evaluations, 153 contained the required deliverables and 27 valid
runs omitted them. The panel mean is 22.24\%.

\section{Domain-Specific Task Sources and Adaptation Records}
\label{app:source-records}

Each task is adapted individually from a public, license-compatible source. Adaptation
freezes the required inputs, maps the original request into an explicit deliverable
contract, and rebuilds or strengthens the evaluator where the upstream check is
insufficient. Shared sources, templates, and near-duplicate workflows receive
leakage-group identifiers for partitioning. Full commit hashes, input checksums,
original requests, licenses, and per-task transformation notes remain inside each
released task package.

\subsection{Presentation}

Table~\ref{tab:presentation-sources} reports the frozen source composition for the 60 presentation tasks.

\begin{table}[H]
    \centering
    \caption{Frozen source inventory for the 60 presentation tasks.}
    \label{tab:presentation-sources}
    \scriptsize
    \begin{tabularx}{\linewidth}{@{}>{\raggedright\arraybackslash}p{0.18\linewidth}c>{\raggedright\arraybackslash}p{0.25\linewidth}>{\raggedright\arraybackslash}X@{}}
        \toprule
        Upstream source & Tasks & License and frozen revision & Upstream task identifiers \\
        \midrule
        PPTArena & 31 & MIT; \nolinkurl{6f26a63} (code), \nolinkurl{20fb890} (data) &
        Cases 001, 003, 004, 014, 015, 016, 021, 023, 025, 027, 028, 029, 030, 031,
        036, 037, 041, 044, 047, 050, 053, 059, 061, 064, 067, 068, 071, 074, 082,
        083, and 093. \\
        OmegaUse-OfficeVal & 17 & Apache-2.0; \nolinkurl{ffbeecb} (code),
        \nolinkurl{cd6ba6d} (data) & OfficeVal 041, 043, 046, 051, 054, 055, 056, 060,
        063, 064, 066, 068, 069, 070, 074, 075, and 076. \\
        AutomationBench & 4 & MIT; \nolinkurl{4a8e106} & Marketing 1115; Finance 4040,
        4044, and 4062. \\
        TheAgentCompany & 4 & MIT; \nolinkurl{98b68ef} &
        \nolinkurl{pm-prepare-meeting-with-customers},
        \nolinkurl{pm-present-gitlab-info-as-ppt},
        \nolinkurl{hr-internal-tooling-slides}, and
        \nolinkurl{ds-stock-analysis-slides}. \\
        Cowork-Bench & 2 & Apache-2.0; \nolinkurl{d943e75} &
        \nolinkurl{sf-department-budget-ppt} and
        \nolinkurl{ppt-clickhouse-executive}. \\
        OfficeBench & 2 & Apache-2.0; \nolinkurl{b978b80} & Tasks 3-2/subtask-0 and
        3-5/subtask-0. \\
        \midrule
        \textbf{Total} & \textbf{60} & & \\
        \bottomrule
    \end{tabularx}
\end{table}

\subsection{Document}

Table~\ref{tab:document-sources} reports the frozen source composition for the 60 document tasks,
which are restricted to tasks whose solver-facing deliverables are native, editable DOCX files.

\begin{table}[H]
    \centering
    \caption{Frozen source inventory for the 60 document tasks.}
    \label{tab:document-sources}
    \scriptsize
    \begin{tabularx}{\linewidth}{@{}>{\raggedright\arraybackslash}p{0.18\linewidth}c>{\raggedright\arraybackslash}p{0.25\linewidth}>{\raggedright\arraybackslash}X@{}}
        \toprule
        Upstream source & Tasks & License and frozen revision & Upstream task identifiers \\
        \midrule
        OfficeBench & 46 & Apache-2.0; \nolinkurl{b978b808} &
        1-12/0, 1-12/1, 1-12/2, 1-13/1, 1-13/0, 1-15/0, 1-15/2, 1-15/1,
        1-16/0, 1-16/1, 1-19/4, 1-19/1, 1-19/0, 1-3/0, 2-18/0, 2-18/1,
        2-18/2, 2-19/0, 2-2/1, 2-2/0, 2-20/0, 2-20/1, 2-20/2, 2-21/0,
        2-22/1, 2-22/0, 2-26/3, 2-26/1, 2-30/0, 2-36/0, 2-37/3, 2-39/3,
        2-39/1, 2-39/0, 2-6/0, 2-7/0, 3-22/0, 3-56/0, 3-58/0, 3-60/0,
        3-63/0, 3-77/0, 3-8/7, 3-82/0, 3-84/0, and 3-91/0. \\
        OmegaUse-OfficeVal & 9 & Apache-2.0; \nolinkurl{ffbeecb8} &
        \nolinkurl{officeval_003}, \nolinkurl{officeval_006}, \nolinkurl{officeval_010},
        \nolinkurl{officeval_018}, \nolinkurl{officeval_022}, \nolinkurl{officeval_029},
        \nolinkurl{officeval_033}, \nolinkurl{officeval_036}, and \nolinkurl{officeval_040}. \\
        llm-docx-editing & 3 & MIT; \nolinkurl{498a4946} &
        D3, F6, and S4. \\
        Cowork-Bench & 1 & Apache-2.0; \nolinkurl{d943e75b} &
        \nolinkurl{rzd-msk-spb-team-trip-gcal-word}. \\
        FORTE & 1 & MIT; \nolinkurl{345c1ec1} &
        \nolinkurl{Legal-020}. \\
        \midrule
        \textbf{Total} & \textbf{60} & & \\
        \bottomrule
    \end{tabularx}
\end{table}

\subsection{Spreadsheet}

Table~\ref{tab:spreadsheet-sources} reports the frozen source composition for the 60 spreadsheet tasks,
each with one native XLSX solver deliverable.

\begin{table}[H]
    \centering
    \caption{Frozen source inventory for the 60 spreadsheet tasks.}
    \label{tab:spreadsheet-sources}
    \scriptsize
    \begin{tabularx}{\linewidth}{@{}>{\raggedright\arraybackslash}p{0.18\linewidth}c>{\raggedright\arraybackslash}p{0.25\linewidth}>{\raggedright\arraybackslash}X@{}}
        \toprule
        Upstream source & Tasks & License and frozen revision & Upstream task identifiers \\
        \midrule
        OfficeBench & 8 & Apache-2.0; \nolinkurl{b978b808} &
        1-6/1, 1-9/2, 1-9/4, 1-5/3, 1-10/3, 1-10/2, 1-9/0, and 1-14/1. \\
        OmegaUse-OfficeVal & 2 & Apache-2.0; \nolinkurl{ffbeecb8} &
        \nolinkurl{officeval_088} and \nolinkurl{officeval_093}. \\
        SpreadsheetBench-2 sample & 19 & CC-BY-SA-4.0; \nolinkurl{d8a14da1} &
        Debugging/01\_03, Debugging/02\_03, Debugging/03\_03,
        Debugging/04\_03, Debugging/08\_03, Debugging/09\_01,
        Financial\_Model/02\_03, Financial\_Model/03\_02,
        Financial\_Model/05\_01, Financial\_Model/07\_02,
        Financial\_Model/12\_01, Financial\_Model/13\_02,
        Financial\_Model/16\_02, Financial\_Model/18\_03,
        Financial\_Model/20\_03, Template/06\_19, Template/14\_09,
        Visualization/Task\_129, and Visualization/Task\_1437004. \\
        Spreadsheet-RL & 12 & CC-BY-SA-4.0; \nolinkurl{6f274cd4} &
        \nolinkurl{excelforum/excel-general/task-job-6f73fa37587dfb5e},
        \nolinkurl{excelforum/excel-general/task-job-1d7e618b48fc03cc},
        \nolinkurl{excelforum/excel-general/task-job-93b8d5ae5b1f85f1},
        \nolinkurl{excelforum/excel-general/task-job-5cd12fc29412660e},
        \nolinkurl{excelforum/excel-general/task-job-2d603b0f50935350},
        \nolinkurl{excelforum/excel-general/task-job-654f78f4eb0df932},
        \nolinkurl{excelforum/excel-general/task-job-2e1e331a1b9dc473},
        \nolinkurl{excelforum/excel-general/task-job-1813a28727a048cb},
        \nolinkurl{excelforum/excel-general/task-job-f59390cb071bb9fd},
        \nolinkurl{excelforum/excel-general/task-job-68594670fd5aafc2},
        \nolinkurl{excelforum/excel-general/task-job-9926a63ec2ac727f}, and
        \nolinkurl{excelforum/excel-general/task-job-b07dec0d38d63663}. \\
        Workspace-Bench-Lite & 4 & Apache-2.0; metadata \nolinkurl{60b08b1c},
        workspaces \nolinkurl{e245d63b} &
        tasks 159, 255, 277, and 386. \\
        Cowork-Bench & 8 & Apache-2.0; \nolinkurl{d943e75b} &
        \nolinkurl{course-enrollment-analytics-dashboard},
        \nolinkurl{financial-audit-cross-system-reconciliation},
        \nolinkurl{insales-inventory-reorder-forecast},
        \nolinkurl{moex-sector-rotation-dashboard},
        \nolinkurl{sf-regional-sales-forecast}, \nolinkurl{sf-sla-breach-audit},
        \nolinkurl{supplier-contract-compliance-audit}, and
        \nolinkurl{inventory-supply-chain-optimization}. \\
        AutomationBench & 4 & MIT; \nolinkurl{4a8e1061} &
        \nolinkurl{finance.annual_budget_prep}, \nolinkurl{finance.cash_flow_forecast},
        \nolinkurl{finance.gl_variance_analysis}, and
        \nolinkurl{finance.vendor_statement_reconciliation}. \\
        TheAgentCompany & 2 & MIT; \nolinkurl{98b68ef8} &
        \nolinkurl{finance-budget-variance} and \nolinkurl{ds-format-excel-sheets}. \\
        FORTE & 1 & MIT; \nolinkurl{345c1ec1} &
        \nolinkurl{Finance-018}. \\
        \midrule
        \textbf{Total} & \textbf{60} & & \\
        \bottomrule
    \end{tabularx}
\end{table}

\subsection{Cross-Office}

Table~\ref{tab:cross-office-sources} reports the source composition for the 60 cross-office tasks.

\begin{table}[H]
    \centering
    \caption{Source inventory for the 60 cross-office tasks.}
    \label{tab:cross-office-sources}
    \scriptsize
    \begin{tabularx}{\linewidth}{@{}>{\raggedright\arraybackslash}p{0.18\linewidth}c>{\raggedright\arraybackslash}p{0.25\linewidth}>{\raggedright\arraybackslash}X@{}}
        \toprule
        Upstream source & Tasks & License and frozen revision & Upstream task identifiers \\
        \midrule
        AutomationBench & 42 & MIT; \nolinkurl{4a8e106} & Sales 1202; Finance 4001;
        HR 5004, 5011, 5057, 5070, 5090, and 5101--5135. \\
        Cowork-Bench & 9 & Apache-2.0; \nolinkurl{d943e75} & Literature review, meal
        planning, program committee, financial audit, portfolio analysis, refund
        analysis, scholarly trends, tax propagation, and student retention tasks. \\
        OfficeBench & 5 & Apache-2.0; \nolinkurl{b978b80} & Tasks 3-6, 3-8, 3-19/3-20,
        3-69, and 3-91. \\
        OmegaUse-OfficeVal & 4 & Apache-2.0; \nolinkurl{ffbeecb} (code),
        \nolinkurl{cd6ba6d} (data) & OfficeVal 059, 078, 080, and 088. \\
        \midrule
        \textbf{Total} & \textbf{60} & & \\
        \bottomrule
    \end{tabularx}
\end{table}

\section{Evolution Split and Information Boundaries}
\label{app:split-boundaries}

Presentation, document, and spreadsheet tasks use a 30/10/20
feedback--selection--final allocation. All 60 cross-office tasks are reserved for
compositional OOD evaluation. Tasks sharing a source document, template, entity, or
near-duplicate workflow are assigned to one split through a common leakage group.

Feedback trajectories and rubric diagnostics are available for updates; selection
exposes a metered aggregate score. Final tasks, answers, and previous static-run
traces are not available to the updater or used to select revisions. Each selected
revision is frozen before final evaluation.

\begin{table}[H]
    \centering
    \caption{Evolution split allocation. Feedback exposes trajectories, artifacts, scores,
    and rubric diagnostics; selection returns a metered aggregate signal; final tasks
    remain inaccessible to evolving agents until the selected harness is frozen.}
    \label{tab:split-boundaries}
    \small
    \begin{tabular*}{\linewidth}{@{\extracolsep{\fill}}lrrrr@{}}
        \toprule
        Modality & Feedback & Selection & Final ID & Final OOD \\
        \midrule
        Presentation & 30 & 10 & 20 & 0 \\
        Document & 30 & 10 & 20 & 0 \\
        Spreadsheet & 30 & 10 & 20 & 0 \\
        Cross-office & 0 & 0 & 0 & 60 \\
        \midrule
        Total & 90 & 30 & 60 & 60 \\
        \bottomrule
    \end{tabular*}
\end{table}

\section{Judge Reliability and Human Audit}
\label{app:judge-reliability}

\subsection{Recorded Checks and Version Boundaries}

The implementation checks in Table~\ref{tab:judge-reliability} cover all 240 tasks
in the current release. Each task uses its frozen Docker
image, reference-solution script, and deterministic judge, with networking
disabled. We retain one result for each of three conditions: the reference
solution's output, an empty output directory, and a corrupt output package.
The corrupt package keeps every reference output filename but replaces every
file's contents with fixed, nonempty invalid bytes. The expected rewards are
1, 0, and 0, respectively, with tolerance $10^{-6}$. This yields 720 reported
checks. Task files and source inputs are checked for changes, and raw judge
results, rewards, image identifiers, and file hashes are retained. The counts
cover this release only.

All 240 reference solutions receive reward 1. For both negative conditions,
seven document tasks retain positive credit: six receive 0.10 for unchanged
inputs, and one receives 0.25 for unchanged inputs and the absence of placeholder
content. Among spreadsheet tasks, 15 empty-output cases and 16 corrupt-output
cases receive positive rewards of 0.05--0.20. Another 18 tasks receive negative
rewards of $-0.20$ to $-0.10$ under both conditions because of existing penalties,
which move the score away from zero without awarding credit. The table measures
exact expected-score behavior: positive residual rewards locate gaps in
invalid-output handling. The audit uses the frozen judges and existing agent
scores.

These checks cover full reference outputs and two severe invalid-output cases;
the human audit below covers judge behaviour on agent artifacts. A separate,
targeted check on seven revised spreadsheet tasks found that constant-formula
substitutions reduced rewards to 0.75--0.79.

\subsection{Sample Frame and Stratification}
\label{app:audit-sampling}

This audit rates \emph{agent outputs} to measure whether the automatic judges
agree with human readers. It is separate from the task review in
Appendix~\ref{app:task-review-interface}, which rates \emph{task packages} and
their reference answers.

The sample frame is every scored candidate output from the model--harness
configurations reported in Table~\ref{tab:static-main-results}, frozen at a
recorded snapshot before sampling: 1{,}440 outputs for the six configurations
completed at the time of sampling. From that frame we draw 240 outputs, 60 per modality.

Within a modality, the draw is stratified by frozen difficulty tier, fifteen
outputs per tier, so that Easy through Expert are equally represented rather than
following the 12/18/18/12 task counts. Within each modality--tier cell, outputs
are stratified again by automatic score into four bands --- exactly zero,
$(0,33]$, $(33,67]$, and $(67,100]$ --- taking at least two outputs from every
non-empty band and allocating the remaining draws in proportion to band
frequency. Zero-score outputs, including missing deliverables, stay in the frame:
they are where a judge error is most consequential. At most two outputs per task
enter the sample, so each modality covers at least 30 distinct tasks, and
harnesses are balanced as evenly as each cell allows.

Raters receive the task instruction, the frozen source inputs, the submitted
artifacts, and the task-specific rubric. They do not receive the harness
identity, the automatic score, the reference solution, or file names that reveal
the producing configuration; presentation order is randomised. Exact-recreation
tasks are the one exception on the reference solution, because the reference
rendering is part of the stated requirement, matching the policy used for the
visual evaluator. Outputs that fail to render because of an infrastructure error
are repaired before review.

\begin{figure}[H]
    \centering
    \includegraphics[width=\linewidth]{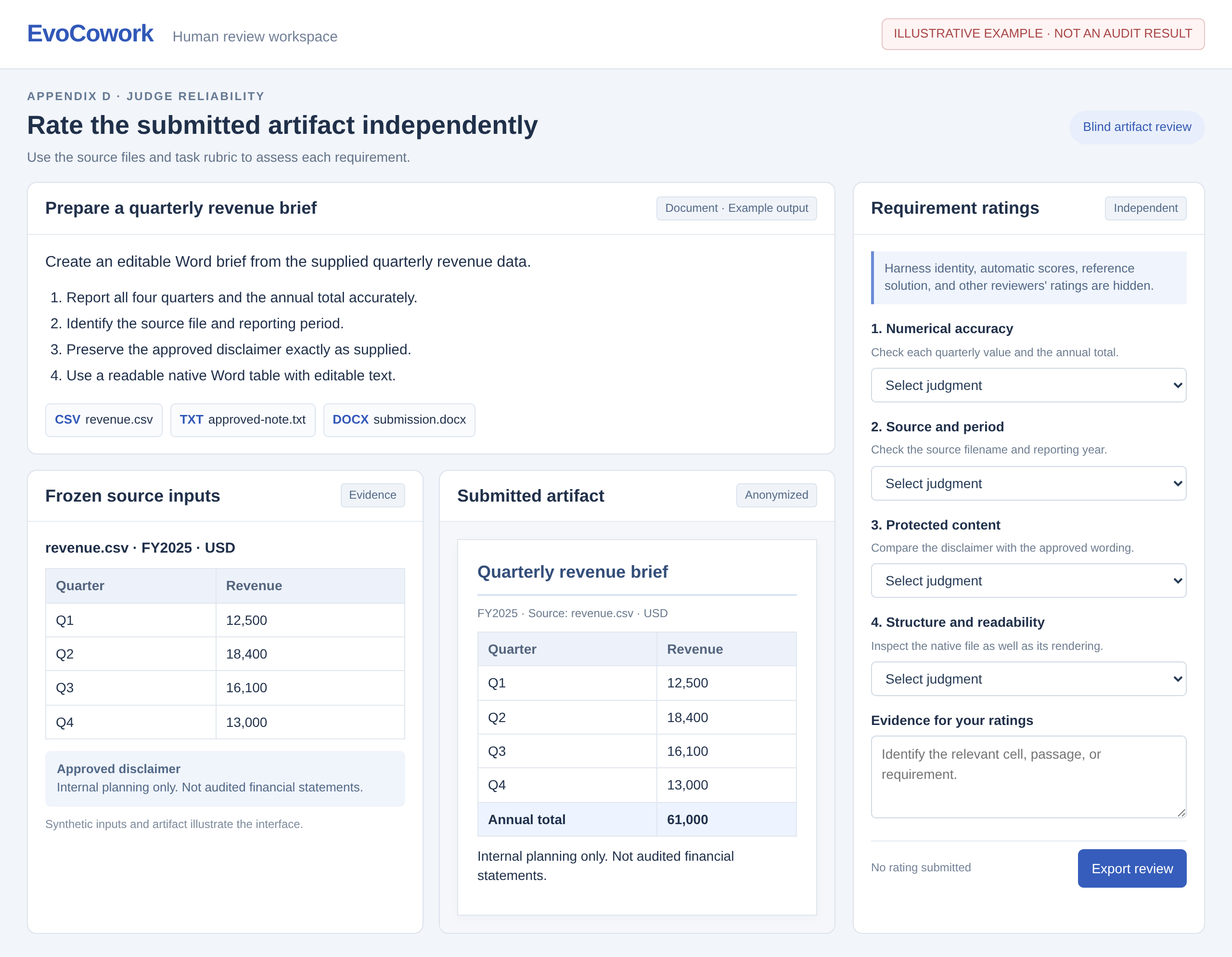}
\end{figure}

\subsection{Rater Panel and Adjudication}
\label{app:audit-raters}

Each modality has a panel of three raters with professional experience in the
corresponding office application. Raters are independent of the material they
audit: none adapted the sampled tasks, implemented or revised their judges, or
took part in the task review of Appendix~\ref{app:task-review-interface}. Each
rater serves on at most two modality panels.

Every sampled output receives two independent ratings, assigned by rotating over
the three pairs within a panel. This yields 120 ratings per modality and 480
ratings in total. Raters record a decision for each rubric requirement on an
ordered scale --- satisfied, partially satisfied, not satisfied --- together with
free-text notes, and they submit before seeing the other rating. Where the two
ratings disagree on a requirement, the third rater on that panel adjudicates and
their decision becomes the human consensus for that requirement. Adjudication
assignments follow from which pair produced the ratings, so each rater covers a
different subset of the sample.

Before the audit proper, all raters score a shared calibration set of eight
outputs, two per modality, and discuss the rubric wording. Calibration ratings
are discarded.

\subsection{Inter-Annotator Agreement and Judge Alignment}

For the ordered rubric decisions, inter-annotator agreement (IAA) is measured
with ordinal Krippendorff's $\alpha$ on the independent, pre-adjudication
ratings, which accommodates raters covering different subsets of the sample.

Judge alignment is measured against the adjudicated consensus as score error.
The human rubric decisions are aggregated into a task score using the same weights
the judge applies, including the 70/30 deterministic--visual combination in
Equation~\ref{eq:empirical-difficulty}, and we report the mean absolute error
(MAE) between the judge score and this human score.

Table~\ref{tab:judge-reliability} reports audited outputs, rater counts,
$\alpha$, and score MAE by modality. Score MAE is in points on the
100-times-reward scale.

\section{Task Review Interface}
\label{app:task-review-interface}

To verify task quality, we randomly sampled tasks from each modality and conducted
a structured human review using a purpose-built web interface. Each reviewer was
presented with the task instruction, input files, and reference answer for one task
at a time. They assessed three aspects: (1) whether the instruction is unambiguous,
(2) whether the reference answer is factually correct and complete, and (3) whether
the reference answer fully satisfies the stated requirements. Reviewers could record
notes and assign an overall verdict of Pass, Flag, or Reject. Flagged and rejected
tasks were revised before inclusion in the released benchmark.

\begin{figure}[H]
    \centering
    \includegraphics[width=\linewidth]{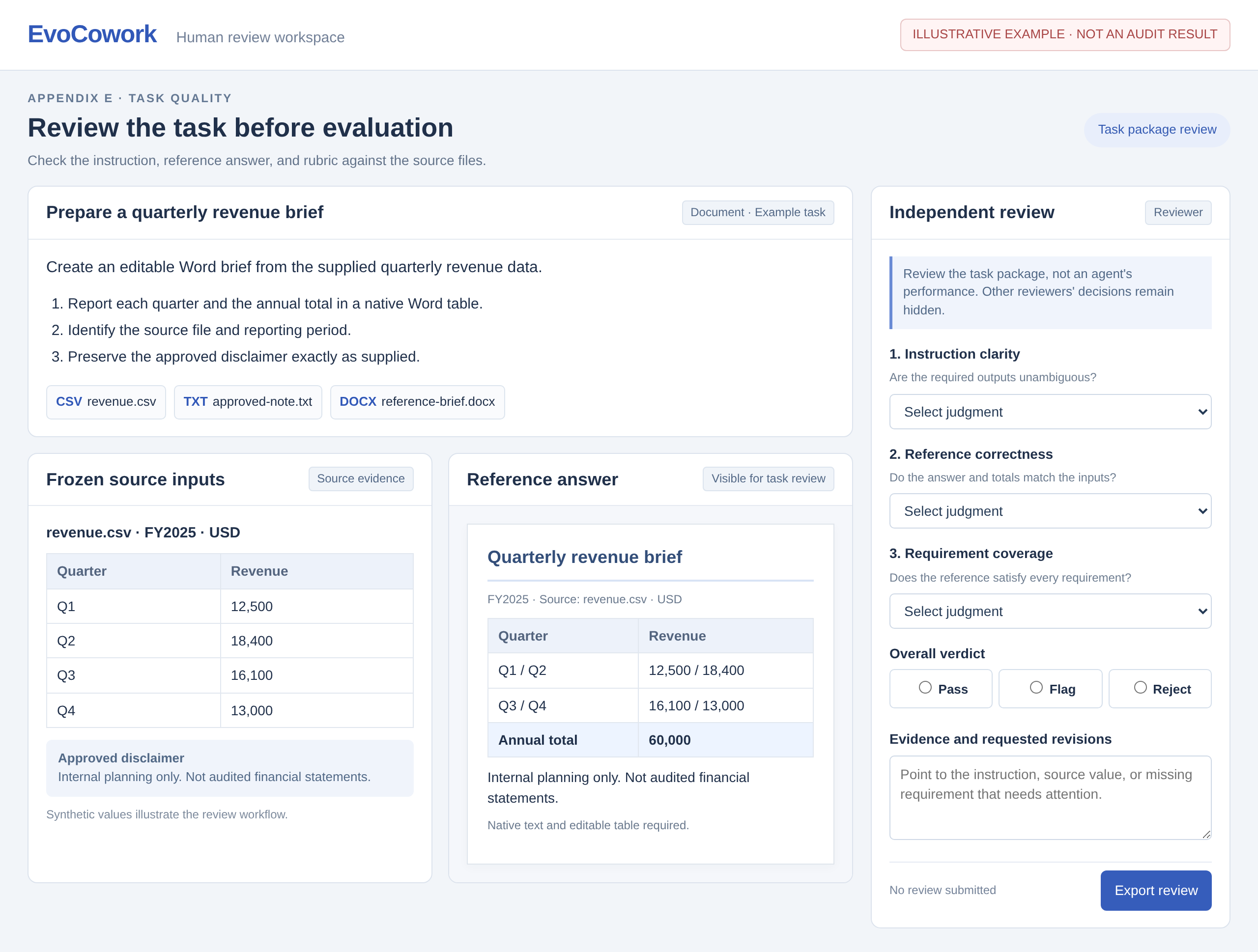}
    \caption{Task review interface used for human quality verification of sampled
    tasks. Reviewers assess instruction clarity, reference answer correctness, and
    requirement coverage. The right panel tracks batch progress and flags tasks
    for revision.}
    \label{fig:task-review-interface}
\end{figure}

Figure~\ref{fig:task-review-interface} shows the review interface. The left panel
displays the task instruction, input files, and reference answer. The right panel
shows the reviewer's assessment form and the progress queue for the current batch.

\subsection{Task Quality Review Records}
\label{app:task-quality-review-records}

The review covers 24 of the 60 tasks in each modality, 96 tasks in total. Tasks
are drawn at random within frozen difficulty tier, six per tier, so that Easy
through Expert contribute equally rather than in proportion to the 12/18/18/12
counts. The draw is stratified by upstream source as well, taking at least one
task from every source that contributes to a modality, so that adaptation
decisions specific to one source cannot escape review. Each modality has a
three-person panel drawn from the authors and from contributors who prepared task
packages for other modalities; nobody reviews a task they adapted themselves.

Every sampled task receives two independent reviews. Reviewers see the
instruction, the frozen inputs, the reference answer, and the task-specific
rubric, and record the three judgments above on an ordered scale --- satisfied,
partially satisfied, not satisfied --- together with a Pass, Flag, or Reject
verdict and free-text notes. They submit before seeing the other review. Where
the two verdicts differ, the third panel member adjudicates and their verdict
becomes the recorded outcome, giving 192 independent reviews plus adjudications
across the sample. Agreement is measured on the independent, pre-adjudication
judgments with ordinal Krippendorff's $\alpha$, alongside the share of tasks on
which both reviewers reach the same verdict.

The review returned 84 Pass, 10 Flag, and
2 Reject verdicts. The two independent reviews agreed on the
verdict for 92 of 96 tasks; the third panel member adjudicated
the remaining 4, and ordinal Krippendorff's $\alpha$ over the
pre-adjudication judgments is 0.86. The 12
flagged or rejected tasks were revised and rechecked before inclusion:
seven for ambiguous instruction wording and five
for reference answers that missed a stated requirement.

Task adaptation, reference-answer construction, and rubric drafting are
AI-assisted and human-checked: an agent proposes the deliverable contract, the
reference solution, and the requirement list from the upstream task, and a person
verifies each against the frozen inputs before the package enters review. The
review above then treats that package as the object under audit. Verdict counts,
the issues they name, and the revision status of each flagged task are kept in the
review log retained with the task package.

\section{Qualitative Failure Analysis}
\label{app:failure-cost}

\paragraph{Failure-rate coverage.}
Table~\ref{tab:failure-analysis} counts a delivery failure when at least one
required file is missing or empty. Static denominators are normally 60 per
modality and 240 overall. All static configurations have 240 task scores.
The Sonnet--Claude Code delivery audit covers all 240 tasks; all 60 PPT outputs
contain every required nonempty file. One GLM--OpenCode Word run has conflicting artifact and judge
records; this run remains unclassified, so the denominators are 59 for Word and
239 overall. Final-test denominators are 20 per single modality, 60 for
cross-office, and 120 overall. Quality columns divide the number of runs with
at least one rule-identified violation by all 240 static or 120 final tasks.
Skipped or unresolved checks are not counted as additional violations, so the
columns report violations the rules identify.

We inspect two static outputs and two paired final-test cases. Each example is
checked against the task files, submitted artifacts, and, where needed, execution
trace. These are illustrative cases rather than a random sample or an estimate
of failure frequency. A failed judge check alone is not treated as sufficient
evidence of an agent error.

\subsection{Human Review of Failure Classifications}
\label{app:failure-classification-review}

This review checks whether the rule-based and agent-assisted classification behind
the quality columns assigns cases to the right category. It is separate from the
audit in Appendix~\ref{app:judge-reliability}, which asks whether the judge scores
match human readers; here the score is taken as given and only the category label
is under review.

The sample frame is every case that the classification flagged in
Table~\ref{tab:failure-analysis}, together with the cases whose requirement text
matched no rule. From that frame we draw 20 cases per quality category, 140 in
total, stratified across the eleven reported configurations so that no configuration
supplies more than a quarter of any category, and stratified within a category
across the four modalities where the check applies. Cases that reached the
agent-assisted pass are sampled at twice their share of the frame, since they carry
the classification decisions least constrained by an explicit rule. Unmatched
checks form a separate stratum of 20 cases, drawn from the same frame.

Each modality panel is the one described in
Appendix~\ref{app:audit-raters}: three reviewers, two independent labels per
case, adjudication by the third where they differ, giving 320 independent labels
plus adjudications. Reviewers see the failing check, the requirement it belongs to,
the submitted artifact, and the relevant source inputs. They do not see the
category the classification assigned, so labels are recorded before any comparison
with the automatic mapping. For each case a reviewer assigns one primary category
from the seven, any additional categories the same case satisfies, or one of two
outcomes outside the taxonomy: the check identifies no error in the artifact, or
the evidence does not settle the question. The second outcome is kept separate from
the first, since an unresolved case is not a confirmed non-error.

Agreement is measured on the independent, pre-adjudication labels with nominal
Krippendorff's $\alpha$ over the primary category, and the adjudicated label is
compared with the automatic mapping to give a per-category agreement rate. Where
adjudication overturns a mapping, both the corrected label and the rule or prompt
that produced the original are recorded, and a category definition is revised only
when the same disagreement recurs across configurations. Corrections that change
any proportion in Table~\ref{tab:failure-analysis} are applied to the table and
noted with the affected cells; the review log retains the case identifiers, the
two independent labels, and the adjudicated outcome.

\begin{figure}[H]
    \centering
    \includegraphics[width=\linewidth]{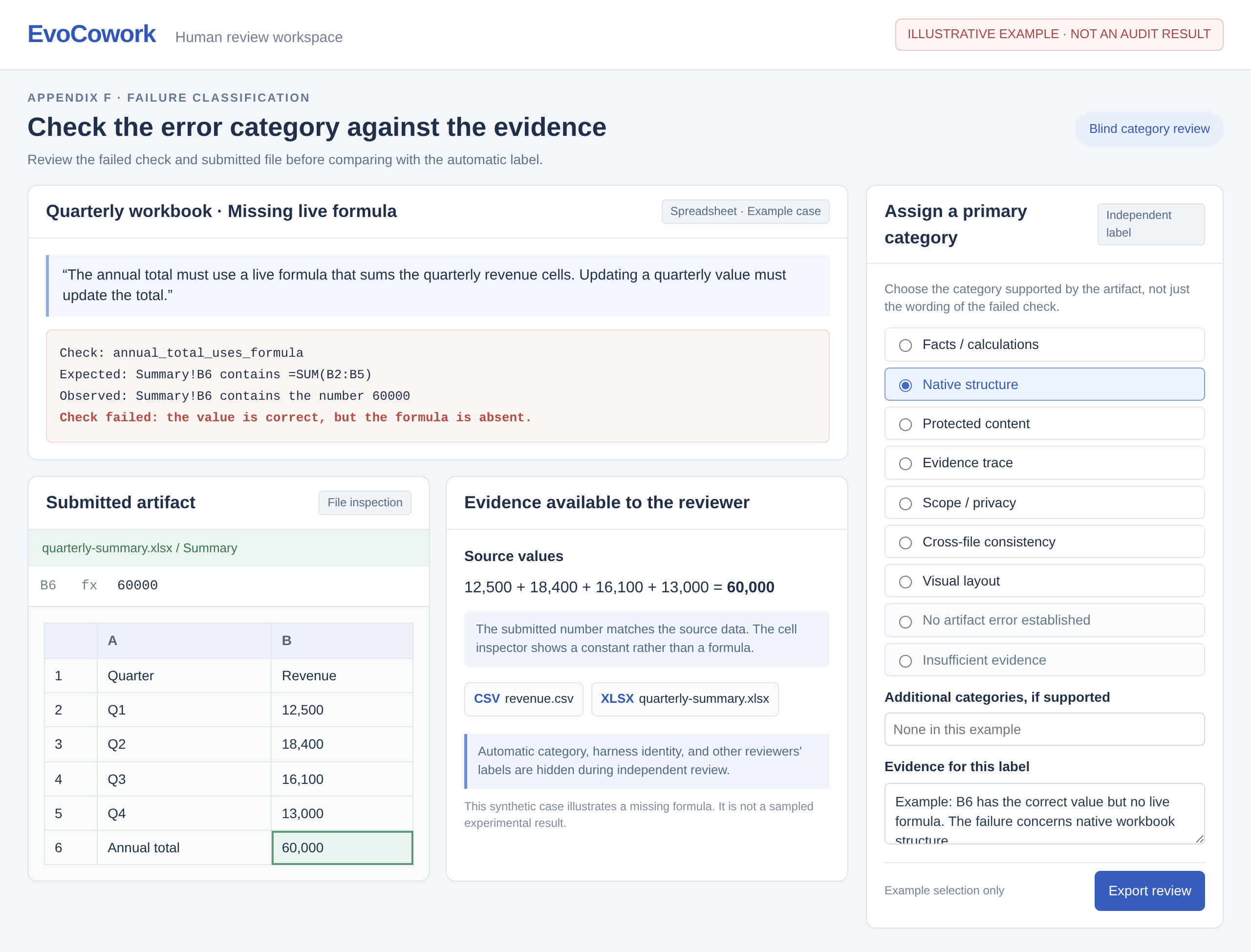}
    \caption{Illustrative interface for reviewing failure classifications.
    The left panels show the requirement, failed check, and artifact evidence;
    the right panel records category labels and supporting evidence.
    The synthetic example shows a correct total stored as a constant instead of
    the required live formula.}
    \label{fig:failure-classification-review}
\end{figure}

Figure~\ref{fig:failure-classification-review} illustrates the review interface
with a synthetic missing-formula example.

\clearpage
\subsection{Static Evaluation Cases}

\begin{figure}[H]
    \centering
    \includegraphics[width=\linewidth]{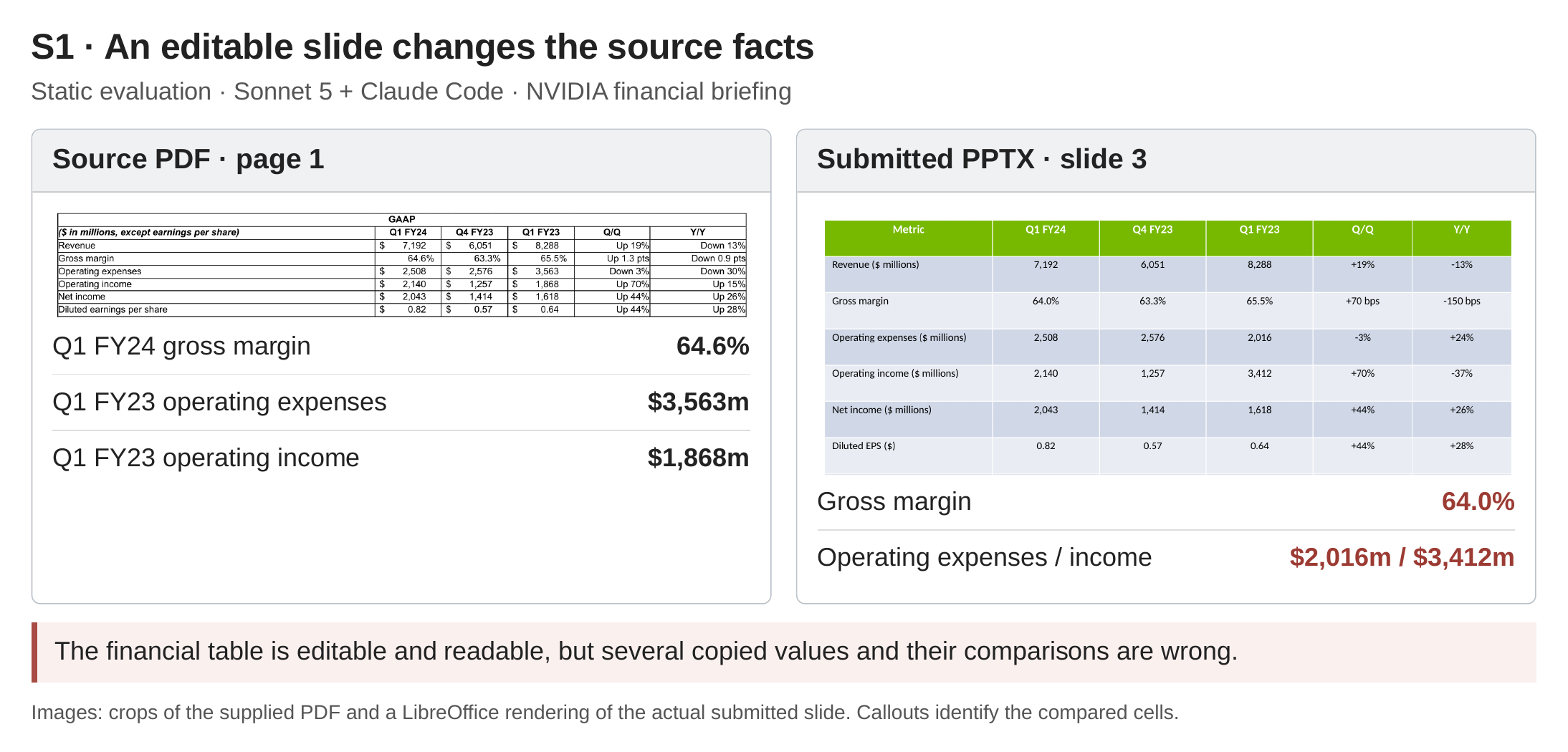}
    \caption{Static case S1: incorrect facts in an editable slide. Crops show
    the supplied PDF and the actual PPTX rendered with LibreOffice; callouts
    identify the mismatched cells.}
    \label{fig:case-static-facts}
\end{figure}

\paragraph{S1: An editable slide changes the source facts.}
The NVIDIA briefing task asks for six editable slides based on a supplied
financial report. Sonnet~5 with Claude Code produces a native table, but slide~3
changes several source values (Figure~\ref{fig:case-static-facts}). Gross margin
becomes 64.0\% instead of 64.6\%; prior-year operating expenses and income are
also incorrect. Checking file existence and editability would miss these
errors. The numbers must also be checked against the source report.

\begin{figure}[H]
    \centering
    \includegraphics[width=\linewidth]{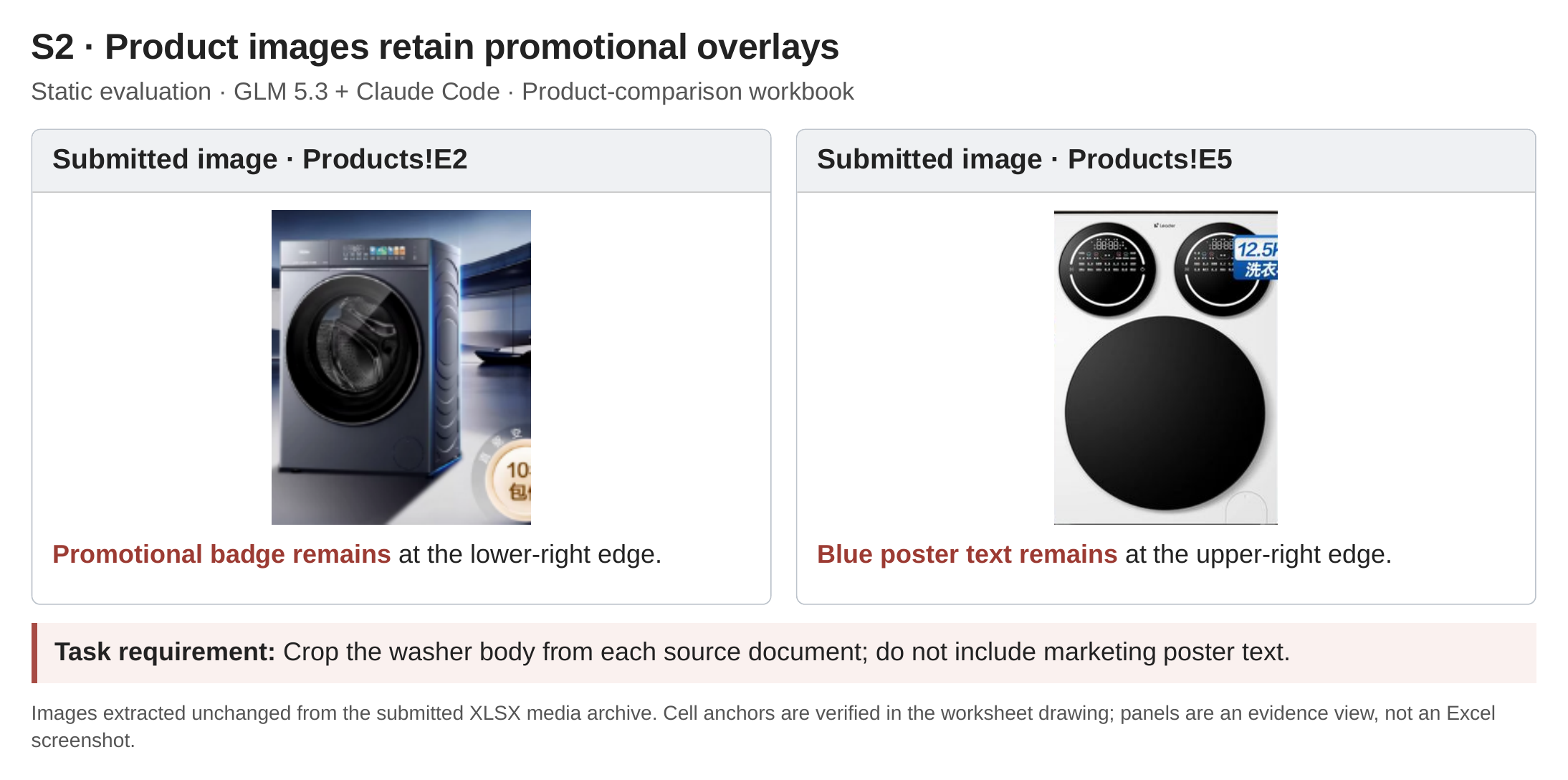}
    \caption{Static case S2: unwanted promotional overlays in product images.
    Both images are extracted unchanged from the submitted XLSX. The worksheet
    drawing anchors them at E2 and E5 of \texttt{Products}.}
    \label{fig:case-static-images}
\end{figure}

\paragraph{S2: Product images retain promotional overlays.}
The product-comparison task asks for a workbook with washer images cropped
from five source documents. It explicitly excludes marketing poster text.
GLM~5.3 with Claude Code embeds five images, but two crops retain promotional
overlays (Figure~\ref{fig:case-static-images}). The image anchored at E2 still
contains part of a badge; E5 retains a blue text banner. File existence and
image count therefore do not establish that the visual requirement is met.
These defects can be checked directly in the embedded images.

\clearpage
\subsection{Evolution Evaluation Cases}

\begin{figure}[H]
    \centering
    \includegraphics[width=\linewidth]{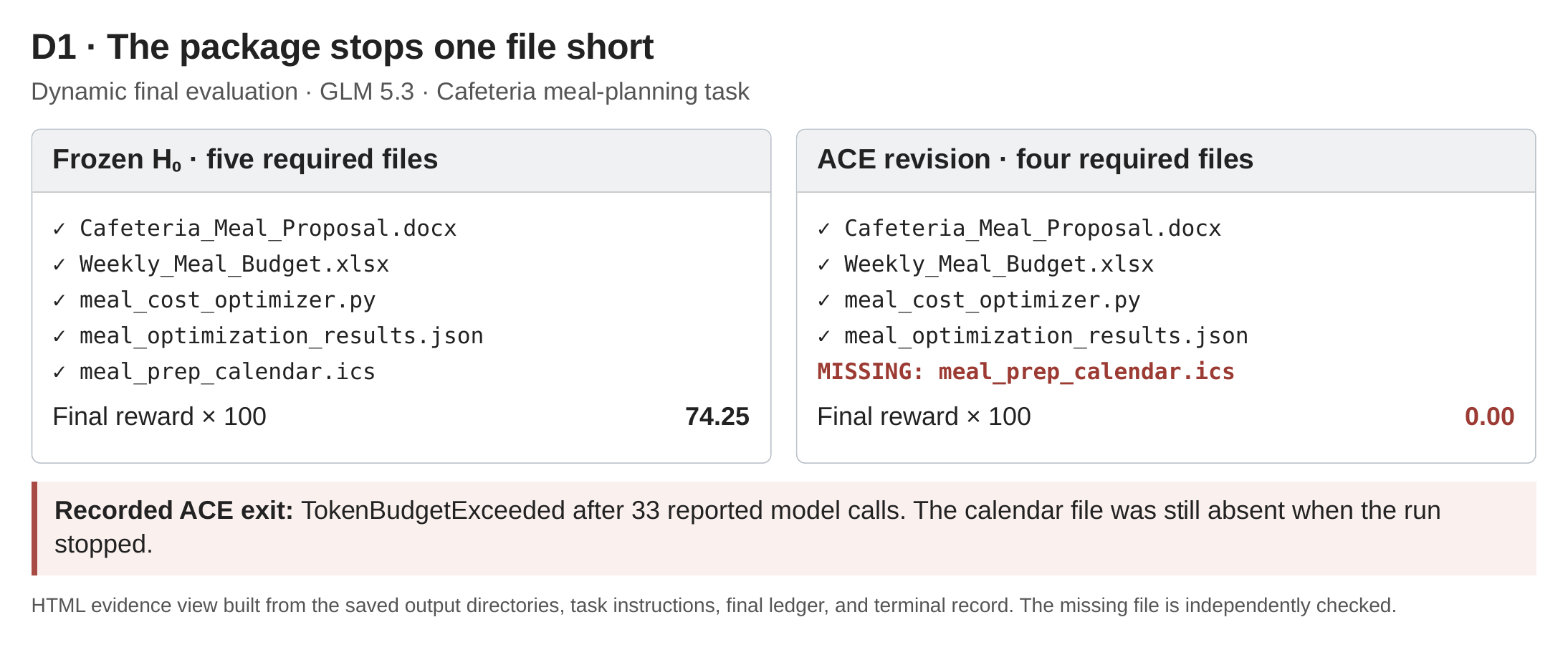}
    \caption{Dynamic case D1: ACE leaves a required calendar undelivered.
    Panels show file presence in the saved output directories, with scores and
    the exit condition from the final ledger and terminal record.}
    \label{fig:case-dynamic-delivery}
\end{figure}

\paragraph{D1: The updated agent stops before completing the package.}
The cafeteria task requires five files. On the same final task, $H_0$ saves all
five, while ACE omits \texttt{meal\_prep\_calendar.ics}
(Figure~\ref{fig:case-dynamic-delivery}). ACE stops at the token-budget guard
after 33 recorded model calls. Both runs reach that guard, but $H_0$ has already
created every required file. The missing calendar fails the delivery hurdle,
giving ACE zero reward. Work on four files does not compensate for the missing
part of this coordinated package.

\begin{figure}[H]
    \centering
    \includegraphics[width=\linewidth]{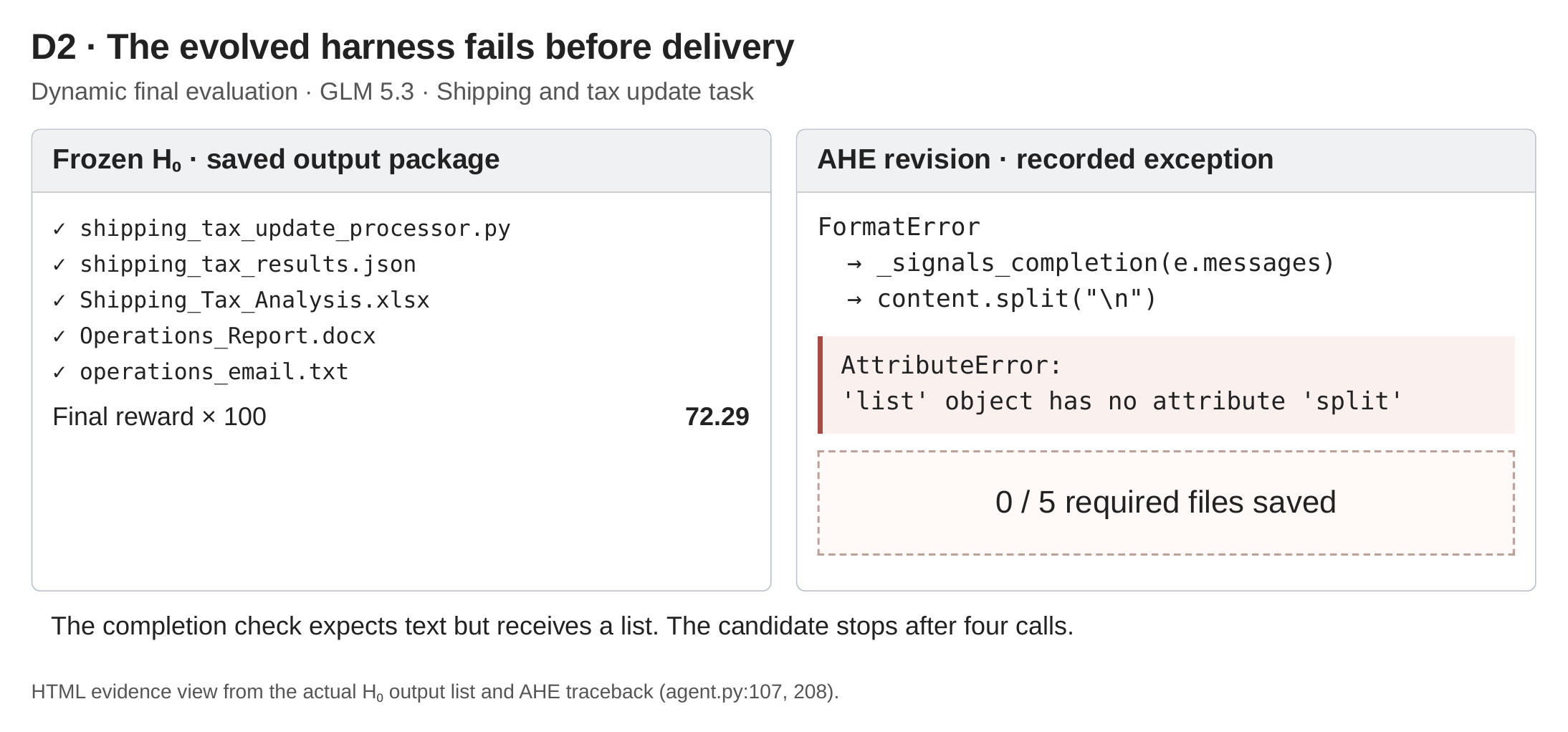}
    \caption{Dynamic case D2: AHE's completion check crashes. Panels use $H_0$'s
    saved file list and the actual AHE traceback; no AHE artifacts were saved.}
    \label{fig:case-dynamic-runtime}
\end{figure}

\paragraph{D2: A revised harness fails during error recovery.}
The shipping-and-tax task requires five coordinated files. $H_0$ produces all
five, while the selected AHE candidate stops after four recorded model calls
without saving any (Figure~\ref{fig:case-dynamic-runtime}). Its trace shows a
format error followed by a crash in the candidate's completion check, which
calls \texttt{split} on a list-valued message. The revised harness fails during
error recovery. It must remain executable when responses take unexpected forms.

\clearpage
\section{Cost and Latency}
\label{app:resource-efficiency}

\subsection{Solver Resources}

\paragraph{Measurement scope.}
Table~\ref{tab:cost-efficiency} reports the solver resources of the selected
task attempts used in the static and final-test results. Token counts sum
reported input and output tokens across calls, including cached input once.
For MSA, we use the selected attempt's gateway records; the other harnesses use
their recorded agent counters. Cache-write and retry coverage vary across these
counters. Execution time runs from agent start to finish,
including tool use and request waits. It excludes container setup, judging,
queueing, and harness search. Failed task outcomes are retained. Missing
measurements are excluded rather than replaced by zero.

\begin{table}[H]
    \centering
    \caption{Solver resources per task in static and evolution evaluations.
    Each pair gives mean reported tokens ($T$, thousands) and mean execution
    time ($t$, minutes). Coverage lists the numbers of available token and time
    records, $(n_T,n_t)$; missing records are excluded from the corresponding
    mean. All includes failed task outcomes. Update generation is excluded.}
    \label{tab:cost-efficiency}
    \begingroup
    \fontsize{8}{9.5}\selectfont
    \setlength{\tabcolsep}{2.1pt}
    \renewcommand{\arraystretch}{1.3}
    \begin{tabularx}{\linewidth}{@{}llc*{8}{>{\raggedleft\arraybackslash}X}>{\columncolor{tablegray}\raggedleft\arraybackslash}X>{\columncolor{tablegray}\raggedleft\arraybackslash}X@{}}
        \toprule
        Model & Harness & Coverage & \multicolumn{2}{c}{PPT} & \multicolumn{2}{c}{Word}
        & \multicolumn{2}{c}{Excel} & \multicolumn{2}{c}{Cross-office}
        & \multicolumn{2}{c}{\cellcolor{tablegray}\textbf{All}} \\
        \cmidrule(lr){4-5}\cmidrule(lr){6-7}\cmidrule(lr){8-9}\cmidrule(lr){10-11}\cmidrule(l){12-13}
        & & $(n_T,n_t)$ & $T$ & $t$ & $T$ & $t$ & $T$ & $t$ & $T$ & $t$ & $\mathbf{T}$ & $\mathbf{t}$ \\
        \midrule
        \graytablerow
        \multicolumn{13}{l}{\textbf{Static evaluation}\quad 240 tasks: 60 per track} \\
         & Claude Code & 239, 239 & 263 & 2.1 & 933 & 8.2 & 904 & 11.0 & 756 & 5.3 & \textbf{716} & \textbf{6.7} \\
         & OpenCode & 240, 240 & 229 & 4.0 & 630 & 7.3 & 334 & 9.8 & 506 & 5.7 & \textbf{425} & \textbf{6.7} \\
         & OpenHands & 228, 240 & 263 & 3.2 & 428 & 7.1 & 885 & 9.1 & 836 & 8.0 & \textbf{610} & \textbf{6.8} \\
        \multirow{-4}{*}{Sonnet 5} & MSA & 240, 240 & 1088 & 10.2 & 740 & 8.5 & 797 & 7.9 & 1389 & 13.8 & \textbf{1003} & \textbf{10.1} \\
        \midrule
         & Claude Code & 240, 240 & 1029 & 10.0 & 739 & 6.9 & 744 & 7.3 & 1233 & 9.6 & \textbf{936} & \textbf{8.5} \\
         & OpenCode & 240, 240 & 646 & 12.5 & 561 & 7.2 & 1097 & 9.5 & 377 & 7.8 & \textbf{670} & \textbf{9.3} \\
         & OpenHands & 221, 240 & 1057 & 9.2 & 683 & 5.8 & 685 & 9.2 & 1604 & 11.5 & \textbf{1028} & \textbf{8.9} \\
        \multirow{-4}{*}{GLM 5.3} & MSA & 240, 240 & 1108 & 10.1 & 720 & 7.1 & 1018 & 9.4 & 1524 & 14.7 & \textbf{1093} & \textbf{10.3} \\
        \midrule
        \graytablerow
        \multicolumn{13}{l}{\textbf{Evolution evaluation}\quad Same final 120 tasks: 20 per single-modality track, 60 cross-office} \\
         & MSA $H_0$ & 120, 120 & 1159 & 9.7 & 748 & 7.7 & 1126 & 10.6 & 1524 & 14.7 & \textbf{1267} & \textbf{12.0} \\
         & MSA + ACE & 120, 120 & 993 & 10.5 & 939 & 8.8 & 1067 & 11.5 & 1461 & 13.3 & \textbf{1230} & \textbf{11.8} \\
        \multirow{-3}{*}{GLM 5.3} & MSA + AHE & 120, 120 & 542 & 7.3 & 588 & 6.4 & 521 & 5.9 & 894 & 9.4 & \textbf{722} & \textbf{8.0} \\
        \bottomrule
    \end{tabularx}
    \endgroup
\end{table}

\paragraph{Observed solver use.}
The reported static configurations average roughly 0.42--1.09 million tokens
and 6.7--10.3 minutes per task. On the shared final set, $H_0$ averages
1.27 million tokens and 12.0 minutes, ACE 1.23 million and 11.8 minutes, and AHE
0.72 million and 8.0 minutes. Early termination, including the candidate crash in
Figure~\ref{fig:case-dynamic-runtime}, reduces both tokens and time while leaving
the requested work incomplete, so resource use is read together with completed
work. Cross-office tasks take longer than each
single-modality track for all three configurations in the final comparison.

\paragraph{Update generation costs.}
Separate updater records contain 2.75 million reported tokens for ACE and
12.66 million for AHE, covering the generation and debugging of updates; solver
calls on feedback and selection tasks are accounted separately. Their budget
accounting charges 2.95 and 13.04 million tokens, respectively, including
conservative charges where usage was unavailable.

\clearpage
\subsection{Updater Cost and Final Gains}
\label{app:cost-gains}

Table~\ref{tab:evolution-cost-gain} places the recorded resources next to the
gains in Table~\ref{tab:evolution-main-results}. Updater tokens include the
formal search history, including unsuccessful updates, rather than only the
steps that produced the selected revision. Final-inference tokens are summed
over the same 120 task attempts used for the reported scores.

\begin{table}[H]
    \centering
    \caption{Recorded resource use and final gains with GLM~5.3 and a shared
    MSA seed. Tokens are reported input plus output, in millions (M). Updater
    use is cumulative over the formal search; final inference covers the 120
    selected task attempts. Gains are score points relative to $H_0$, computed
    before rounding. $H_0$ is frozen and runs no updater. These stages exclude
    feedback/selection solver calls and judging, so their sum is not a complete
    experiment cost.}
    \label{tab:evolution-cost-gain}
    \small
    \setlength{\tabcolsep}{4pt}
    \renewcommand{\arraystretch}{1.2}
    \begin{tabular*}{\linewidth}{@{\extracolsep{\fill}}lrrrrr@{}}
        \toprule
        & \multicolumn{2}{c}{Reported tokens (M)} & \multicolumn{3}{c}{Final gain over $H_0$} \\
        \cmidrule(lr){2-3}\cmidrule(l){4-6}
        Method & Updater & Final inference & In-domain & Cross-office & Overall \\
        \midrule
        \graytablerow
        Frozen $H_0$ & n/a & 152.07 & 0.00 & 0.00 & 0.00 \\
        ACE & 2.75 & 147.64 & \gainpoints{4.90} & \gainpoints{0.03} & \gainpoints{2.46} \\
        AHE & 12.66 & 86.65 & \gainpoints{2.64} & \losspoints{1.72} & \gainpoints{0.46} \\
        \bottomrule
    \end{tabular*}
\end{table}

In these runs, AHE uses 4.60 times as many reported updater tokens as ACE,
but gains 0.46 overall points compared with ACE's 2.46. Their in-domain gains
are 2.64 and 4.90 points; cross-office gains are $-1.72$ and $+0.03$ points.
More updater computation therefore does not correspond to a larger gain in
this comparison. AHE uses fewer tokens during final inference, but the
candidate failure in Figure~\ref{fig:case-dynamic-runtime} shows why low
inference use must be read together with completed work.

\section{Case Study of Harness Revisions}
\label{app:harness-revisions}

\begin{figure}[t]
    \centering
    \includegraphics[width=\linewidth]{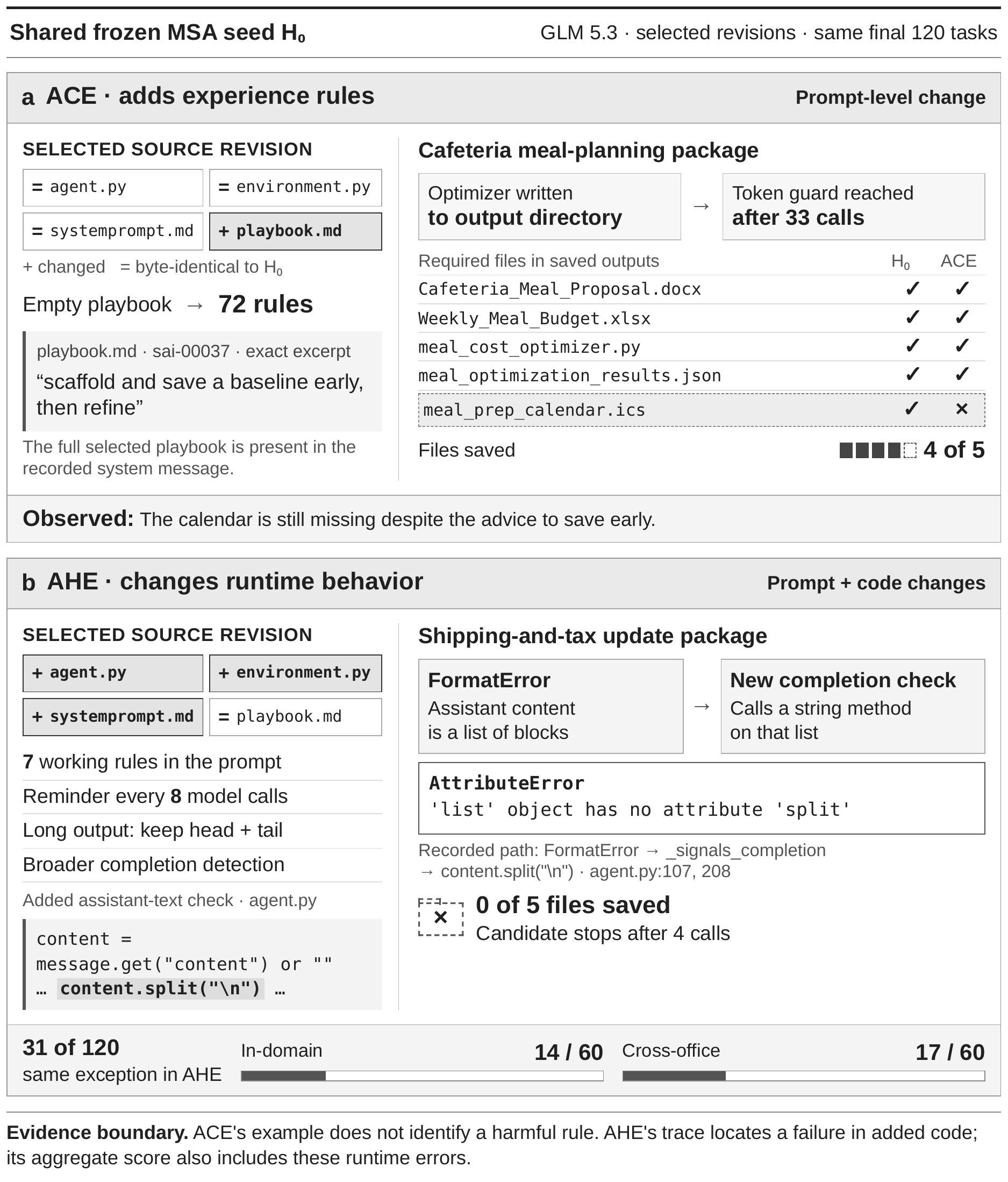}
    \caption{Selected harness changes and recorded outcomes. (a) ACE adds a
    playbook but still misses a required calendar. (b) AHE's added completion
    check crashes on list-valued content. Panels reorganize verified source,
    tool actions, file inventories, and terminal records; the bars count the
    same exception across all AHE final attempts.}
    \label{fig:harness-revisions}
\end{figure}

We compare the exact frozen revisions used for the final results with their
shared initial harness. File hashes match the candidate IDs in all 120 final
records for each method. Figure~\ref{fig:harness-revisions} connects the changes
to two recorded task outcomes. Table~\ref{tab:harness-revisions} gives details
of the source changes on the next page.

\subsection{Revision Details}

\begin{table}[H]
    \centering
    \caption{Changes in the selected ACE and AHE harnesses relative to $H_0$.
    Entries describe source behavior, not separately measured contributions to
    score. The model, API, and inference budgets remain outside these revisions.}
    \label{tab:harness-revisions}
    \footnotesize
    \setlength{\tabcolsep}{4pt}
    \renewcommand{\arraystretch}{1.2}
    \begin{tabularx}{\linewidth}{@{}p{0.16\linewidth}XXX@{}}
        \toprule
        \rowcolor{tablegray}
        Component & Frozen $H_0$ & ACE & AHE \\
        \midrule
        Changed files & Initial four-file snapshot
        & \texttt{playbook.md}
        & \texttt{agent.py}, \texttt{environment.py}, \texttt{systemprompt.md} \\
        \addlinespace[3pt]
        Instructions & Base office prompt; empty playbook
        & Adds 72 experience rules; base prompt unchanged
        & Adds seven working rules to the prompt; playbook stays empty \\
        \addlinespace[3pt]
        Progress cues & No periodic reminder
        & Unchanged
        & Adds a reminder every eight model calls \\
        \addlinespace[3pt]
        Output handling & No truncation inside \texttt{environment.py}
        & Unchanged
        & Above 12,000 characters, keeps the first 8,000 and last 3,500 plus a notice \\
        \addlinespace[3pt]
        Completion & Signal must be the first output line of a successful command
        & Unchanged
        & Accepts any standalone output line; adds an assistant-text check after a format error \\
        \bottomrule
    \end{tabularx}
\end{table}

\paragraph{ACE: experience rules without changes to the execution loop.}
ACE changes only \texttt{playbook.md}, from an empty file to 72 entries
(48,917 bytes). The agent loop, environment code, and base system prompt match
$H_0$ byte for byte. The entries cover native Office objects, exact labels,
checks against source files, and saving deliverables before the budget runs out.
Some explicitly target how checks extract text, including duplicating key
figures in tables and text boxes. These rules could improve extractability;
they also make exact wording and the scope of each rule important.

The cafeteria run in Figure~\ref{fig:case-dynamic-delivery} receives this full
playbook. Rules \texttt{sai-00037}, \texttt{code-00039}, and
\texttt{misc-00068} advise creating files early and reserving time to submit.
The trace records the optimizer being written directly to the output directory,
but the run still ends with four of five required files: guidance in the
playbook does not by itself produce a complete package.

\paragraph{AHE: changes to reminders, tool output, and completion handling.}
AHE leaves the playbook empty and changes the other three files. Its new
instructions emphasize early file creation, short commands, and concrete
checks before submission. The agent adds a reminder every eight model calls,
and the environment shortens long command output while retaining its beginning
and end. These changes target prolonged inspection; truncation trades that
against hiding information in the middle of an output.

A separate change broadens completion detection. The new format-error handler
checks assistant text for the completion signal, using
\texttt{content.split("\textbackslash n")}. In the shipping-and-tax run
(Figure~\ref{fig:case-dynamic-runtime}), the response content is a list of
blocks. That new check raises \texttt{AttributeError} after four recorded
model calls, before any required file is saved. The source diff and traceback
locate this failure in an added completion check. The same exception is
recorded in 31 of 120 AHE final attempts: 17 cross-office and 14 in-domain.
No such exception is recorded for $H_0$ or ACE. These crashes do not all receive
zero reward, since some runs save artifacts before failing. This repeated
compatibility problem limits how much of AHE's aggregate result can be
attributed to the intended update strategy.

\subsection{Crash-Excluded Diagnostic for AHE}

Table~\ref{tab:ahe-crash-split} splits AHE's 120 final-test runs by whether
the added completion check crashed. The decline in
Table~\ref{tab:evolution-main-results} is concentrated in the crashing runs,
which average 6.04 overall and 0.45 on cross-office tasks, against 48.75 and
37.44 on the runs that do not trigger the crash. On those runs, AHE scores
above $H_0$ on the same tasks by 6.41 points overall and 4.87 points
cross-office. The grouping is defined after seeing the failures, so the
clean-subset difference is a diagnostic of the crash's contribution rather
than a repaired-AHE result; the reported scores keep every run, crashed or
not.

\begin{table}[H]
    \centering
    \caption{Diagnostic split of AHE's final-test runs by whether the added
    completion check crashed. Scores are 100$\times$ mean reward on each task
    group; the clean $\Delta$ compares AHE with $H_0$ on the runs that do not
    trigger the crash. The split is a post-hoc diagnostic by run-log status and
    does not represent a repaired AHE.}
    \label{tab:ahe-crash-split}
    \small
    \setlength{\tabcolsep}{4pt}
    \renewcommand{\arraystretch}{1.15}
    \begin{tabular*}{\linewidth}{@{\extracolsep{\fill}}lrrrrrrr@{}}
        \toprule
        Task group & Runs & Crashes & Share
        & \shortstack{AHE\\crashed} & \shortstack{AHE\\clean}
        & \shortstack{$H_0$\\same tasks} & \shortstack{Clean\\$\Delta$} \\
        \midrule
        \graytablerow
        All final tasks & 120 & 31 & 25.8\% & 6.04 & 48.75 & 42.34 & \gainpoints{6.41} \\
        \midrule
        In-domain & 60 & 14 & 23.3\% & 12.84 & 59.32 & 51.48 & \gainpoints{7.84} \\
        \quad PPT & 20 & 5 & 25.0\% & 13.91 & 66.19 & 59.17 & \gainpoints{7.02} \\
        \quad Word & 20 & 3 & 15.0\% & 30.05 & 59.03 & 48.02 & \gainpoints{11.02} \\
        \quad Excel & 20 & 6 & 30.0\% & 3.33 & 52.31 & 47.45 & \gainpoints{4.86} \\
        \midrule
        Cross-office & 60 & 17 & 28.3\% & 0.45 & 37.44 & 32.57 & \gainpoints{4.87} \\
        \bottomrule
    \end{tabular*}
\end{table}

\section{Reproducibility and Experimental Configurations}
\label{app:experimental-configurations}

All-task static diagnostics and held-out-model difficulty validation use fixed
analysis sets. Each configuration is run once per task, so the scores carry no
repeat-level variance; uncertainty estimates over leakage groups and repeated
runs are not part of this release. Valid agent outcomes receive task rewards.
Table~\ref{tab:experimental-configurations} records the model and harness
versions and the solver limits recovered from the saved run configurations.

\subsection{Static Evaluation Details}
\label{app:static-evaluation-details}

The Sonnet~5 panel uses Claude Code~2.1.251, OpenCode~1.4.14, and OpenHands
SDK~1.44.1. Each task contributes one selected completion per harness. Document
and spreadsheet results include the pilot runs and reused scores described in
Appendix~\ref{app:difficulty-taxonomies}.

GLM~5.3 uses the same harnesses and fixed Sonnet~5 difficulty labels. Claude Code,
OpenCode, and OpenHands each have all 240 task scores, including visual scoring
where required. Missing deliverables receive the original judge's failure score,
and native spreadsheet penalties are retained.
The frozen MSA~2.4.6 run with GLM~5.3 has valid scores for all 240 tasks,
including visual scoring where required. Its All, Easy, Medium, Hard, and Expert
rows use the same fixed difficulty labels; each modality has 60, 12, 18, 18,
and 12 tasks in these respective groups. The independent Sonnet~5--MSA run
uses the same frozen harness, task set, and scoring runtime and completes all
240 tasks, scoring 38.08 overall. Its difficulty groups use the same fixed labels.
Thirty tasks needed retries after provider failures; the first valid outcome is retained.
The per-attempt solver limits are unchanged. Model settings and resource
limits are recorded with each configuration.

\begin{table}[H]
    \centering
    \caption{Settings used for the static model--harness comparisons.
    Token caps are cumulative per solver attempt; time limits are in seconds.
    Not set means that the benchmark runner imposes no separate cap.}
    \label{tab:experimental-configurations}
    \footnotesize
    \setlength{\tabcolsep}{3pt}
    \renewcommand{\arraystretch}{1.15}
    \begin{tabular*}{\linewidth}{@{\extracolsep{\fill}}lllrrr@{}}
        \toprule
        Model & Harness version & \shortstack[l]{Call/iteration\\limit}
        & \shortstack{Token\\cap} & \shortstack{Time\\cap (s)} & \shortstack{Runs/\\task} \\
        \midrule
         & Claude Code 2.1.251 & 80 turns & Not set & 600--3{,}600 & 1 \\
         & OpenCode 1.4.14 & Not set & Not set & 600--3{,}600 & 1 \\
         & OpenHands SDK 1.44.1 & 80 iterations & Not set & 600--3{,}600 & 1 \\
        \multirow{-4}{*}{Sonnet 5} & MSA 2.4.6 & 80 API calls & 2{,}000{,}000 & 600--3{,}600 & 1 \\
        \midrule
         & Claude Code 2.1.251 & 80 turns & Not set & 600--3{,}600 & 1 \\
         & OpenCode 1.4.14 & Not set & Not set & 600--3{,}600 & 1 \\
         & OpenHands SDK 1.44.1 & 80 iterations & Not set & 600--3{,}600 & 1 \\
        \multirow{-4}{*}{GLM 5.3} & MSA 2.4.6 & 80 API calls & 2{,}000{,}000 & 600--3{,}600 & 1 \\
        \midrule
        GPT6 Astra & MSA 2.4.6 & 80 API calls & 2{,}000{,}000 & 600--3{,}600 & 1 \\
        Kimi K3 & MSA 2.4.6 & 80 API calls & 2{,}000{,}000 & 600--3{,}600 & 1 \\
        Qwen & MSA 2.4.6 & 80 API calls & 2{,}000{,}000 & 600--3{,}600 & 1 \\
        \bottomrule
    \end{tabular*}
\end{table}

Table~\ref{tab:experimental-configurations} reports the settings of the saved
static runs; the last three rows cover the Expert-tier comparison in
Figure~\ref{fig:expert-model-comparison}, which runs on the same MSA settings.
Claude Code uses an 80-turn limit, OpenHands allows 80 iterations,
and MSA allows 80 model calls including request retries; these counters measure
different events. The runner sets no separate OpenCode step cap and no cumulative
token cap for Claude Code, OpenCode, or OpenHands. MSA has a two-million-token
cap per attempt. Client and provider limits still apply where the table says
Not set. All configurations use the solver timeout in each task's
\texttt{task.toml}, ranging from 600 to 3{,}600 seconds, separately from setup and
judging. Each task contributes one retained outcome; infrastructure retries do
not count as independent repeats.

\subsection{Evolution Evaluation Details}
\label{app:evolution-evaluation-details}

The initial harness is a frozen mini-SWE-agent (MSA)~2.4.6 adapted to office
tasks, with GLM~5.3 as the solver and updater. ACE updates the context playbook;
AHE may revise the MSA prompt, tools, and agent loop. Both start from the same
harness snapshot. The task manifest and source-overlap audit define the feedback,
selection, and final partitions (Appendix~\ref{app:split-boundaries}).

ACE and AHE each have a cumulative updater budget of 100 million tokens, covering
all analysis roles and retries. Both permit at most three candidate submissions,
four selection queries including $H_0$, and 24 hours of search. Token consumption
accumulates across resumed runs against the same limit.
Search and inference costs are reported separately. Final inference uses the same
per-task budget and scoring rules for all three rows, and the paired baseline uses
the same final task IDs, task versions, and evaluation settings.

For reporting, we evaluate the highest-scoring evolved revision on selection
without replacing it with $H_0$ when it scores lower. This reporting rule was
added after inspecting the AHE selection result, before testing that revision
on the final tasks. The original gate decision and its reused $H_0$ test records
are retained separately. No additional selection query or harness update is
made for this evaluation; the revised harness is frozen before final testing.
Only one AHE revision completed selection evaluation. A subsequent feedback
pass reached 58 of 90 tasks before the 24-hour search limit and did not produce
another evaluated revision.

\section{Conclusion and Limitations}
\label{app:discussion-limitations}

\subsection{Conclusion}
\label{sec:conclusion}

\textsc{EvoCowork} evaluates whether changes learned from office-task feedback
help agents complete new work, including workflows that require several files
to agree. Its 240 tasks combine artifact-level scoring with separate feedback,
selection, and final sets, so improvements can be tested beyond the tasks used
to develop a harness. In our GLM~5.3 experiments with a shared frozen MSA seed,
ACE and AHE improve the in-domain average, while cross-office gains are near
zero for ACE and negative for AHE. These results show why office-agent
self-improvement needs a separate test of cross-office transfer: a higher
overall score can hide little improvement or a decline on those workflows.
A blinded audit of 60 outputs per modality places the judges within 3.8 to 9.6
score points of the human consensus, with rater agreement from 0.969 to 0.994,
and the modality ordering differs from the one the deterministic checks produce.
The benchmark provides a common setting for comparing harness revisions and
examining whether their gains extend to new tasks and justify their cost.

\subsection{Limitations}
\label{app:limitations}

The benchmark measures offline office deliverables in isolated task environments,
so its scores describe artifact quality under replayable conditions rather than
live application state, unrestricted web access, or deployment with real business
consequences. Source adaptation improves replayability and can change the
interaction required by the upstream task, so cross-suite comparisons use the
adaptation records in Appendix~\ref{app:source-records}.

Empirical difficulty is conditional on the calibration panel and current judges.
Rank bins impose an ordering on that panel, and the document Hard--Expert boundary
falls inside a block of tied zero scores, resolved by tie-breaking. The GLM~5.3
configurations in Table~\ref{tab:static-main-results} test the fixed labels on a
held-out model. Visual scoring may introduce evaluator-specific preferences, while
deterministic checks are sensitive to artifact structure and exact formulas. The
reported comparisons cover the frozen configurations tested here.

\end{document}